%% file: main.tex
\documentclass{article}

\usepackage{iclr2027_conference,times}
\input{math_commands.tex}

\usepackage{booktabs}
\usepackage{graphicx}
\usepackage{amssymb}
\usepackage{flafter}
\usepackage{placeins}
\usepackage{microtype}
\usepackage{hyperref}
\usepackage{url}
\hypersetup{hidelinks}

\title{The Answer Is Not the State:\\
Trajectory-Level Corrigibility in Diffusion Language Models}

\author{Junchi Liao \quad Wentao Li}

\iclrfinalcopy
\hypersetup{pdftitle={The Answer Is Not the State: Trajectory-Level Corrigibility in Diffusion Language Models},pdfauthor={Junchi Liao, Wentao Li}}

\begin{document}

\maketitle
\lhead{}

\begin{abstract}
\input{manuscript/00_abstract.tex}
\end{abstract}

\input{manuscript/01_introduction.tex}
\input{manuscript/02_method.tex}
\input{manuscript/03_experiments.tex}

\input{manuscript/04_conclusion.tex}
\input{manuscript/05_statements.tex}

\bibliography{bibliography/references}
\bibliographystyle{iclr2027_conference}

\clearpage
\appendix
\input{appendix/A_related_work.tex}
\input{appendix/B_limitations.tex}
\input{appendix/C_experimental_details.tex}
\input{appendix/D_additional_results.tex}
\clearpage
\input{appendix/E_prompt_restart.tex}

\end{document}

%% file: math_commands.tex
\usepackage{amsmath,amsfonts,bm}

\def\eqref#1{equation~\ref{#1}}

\def\1{\bm{1}}

\DeclareMathAlphabet{\mathsfit}{\encodingdefault}{\sfdefault}{m}{sl}
\SetMathAlphabet{\mathsfit}{bold}{\encodingdefault}{\sfdefault}{bx}{n}



%% file: manuscript/00_abstract.tex
Diffusion language models generate tokens at many positions on a shared canvas, making unfinished responses appear revisable. We find a counterexample: after the final answer is reopened, previously generated rationale can reconstruct the old decision despite newly available evidence. We isolate this effect by comparing reopening the answer alone with reopening both rationale and answer, and by exchanging rationales between paired trajectories. On a 150-question source-disjoint benchmark, reopening both regions makes decisions follow the evidence currently present---whether correct or deliberately corrupted---across Dream, LLaDA, and iLLaDA. On a new 100-question free-form short-answer evaluation, the same effect remains strong in LLaDA and iLLaDA but is negligible in Dream. Coherent transplanted rationales steer rebuilt decisions, while token-level attribution improves which positions to reopen without producing a compact repair. Corrigibility therefore depends on retained rationale state, not answer openness alone.

%% file: manuscript/01_introduction.tex
\section{Introduction}
\label{sec:introduction}

Diffusion language models turn generation order into a policy. Rather than emit tokens in a fixed left-to-right sequence, they repeatedly predict across a shared masked canvas. The model proposes token values, and a commitment policy decides which positions become visible state~\citep{austin2021d3pm,sahoo2024mdlm,nie2025llada,ye2025dream}. This flexibility enables arbitrary-order and blockwise generation, semantic planning, and early answer commitment~\citep{arriola2025block,israel2026planned,li2026prophet}, but it also determines when generated text begins to condition later predictions. If a retrieved passage, tool result, or verification signal arrives after part of the canvas has committed, can the model still revise its decision? An unfinished canvas is not necessarily a revisable decision state.

Under monotone decoding, committed positions remain visible to every later denoising step. We call a trajectory \emph{corrigible} when the evidence currently present can still control its decision after such commitments have formed. In our timing experiments with Dream and LLaDA, keeping the answer masked does not guarantee that later evidence can redirect the decision: other visible commitments can already constrain what the answer becomes. Under the monotone decoder studied here, answer openness is necessary for revision but does not describe the full decision state. Prior work shows that intermediate tokens can steer denoising trajectories~\citep{li2026diffuguard,yamabe2026priming,jeung2026a2d} (see Appendix~\ref{app:related-work}); we ask which retained visible content prevents newly arrived evidence from redirecting the decision.

\begin{figure}[t]
    \centering
    \includegraphics[width=\linewidth]{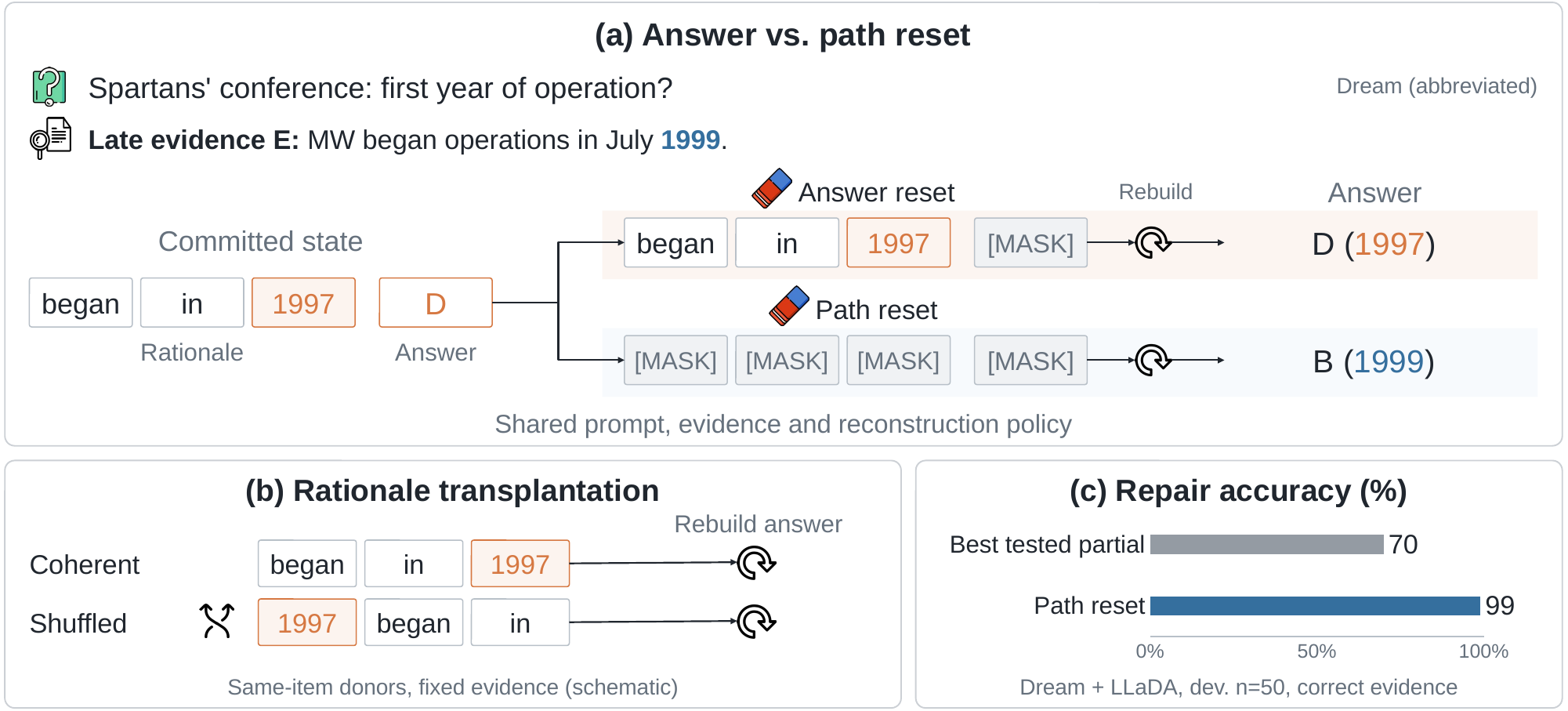}
    \caption{\textbf{Visible rationale is causal trajectory state.} (a) In a Dream example, answer reset reconstructs the prior decision despite late correct evidence, whereas path reset changes it. (b) A separate same-item transplantation compares coherent donor rationale with a shuffle of the same tokens under fixed recipient evidence. (c) The best tested partial reset remains below the full-path endpoint. Case text is abbreviated and token strips are schematic.}
    \label{fig:overview}
\end{figure}

The answer is not the state. Committed rationale tokens remain on the canvas and condition later predictions; here, \emph{state} refers to these visible tokens, not hidden activations. We compare reopening the answer alone (\emph{answer reset}) with a current-canvas reset of both rationale and answer (\emph{path reset}). The prompt, current evidence, intervention time, and reconstruction policy remain fixed, so their contrast isolates the retained rationale. We then exchange exact rationales between paired trajectories and compare each coherent donor with a shuffle of the same tokens. Figure~\ref{fig:overview} summarizes these interventions.

\FloatBarrier

The HotpotQA experiments show a strong path-level effect for constrained decisions and a model-dependent effect for free-form short answers. On the 150-item source-disjoint holdout, reopening both rationale and answer gains $49.6$ points over reopening the answer alone after subtracting the corresponding change with absent evidence. Correct and controlled corrupt evidence drive decisions toward opposite endpoints. On the new 100-item free-form holdout, the corresponding gains are $+46$ points in LLaDA, $+63$ in iLLaDA, and $+1$ in Dream.

Two complementary interventions reveal how retained rationale is organized. Replacing the rationale with one built under opposite evidence moves the rebuilt answer $28.3$ points further toward the donor-associated answer than shuffling the same tokens. Ranking rationale positions by how much their removal changes the decision improves which positions to reopen, yet the best tested partial reset reaches 70\% accuracy versus 99\% after reopening the full path.

\paragraph{Contributions.}
We make three contributions:
\begin{enumerate}
    \item We define \emph{trajectory-level corrigibility}, separating answer openness from trajectory revisability.
    \item We identify committed rationale as persistent visible decision state and show that reopening it transfers control to current evidence.
    \item We establish an attribution--repair gap: token prioritization improves without yielding a compact sufficient repair.
\end{enumerate}

%% file: manuscript/02_method.tex
\section{Method}
\label{sec:method}

We identify which visible commitments determine whether late evidence can revise a decision. Matched reset comparisons begin from trajectories that share the same pre-arrival state. Transplant comparisons instead hold the recipient prompt, evidence, intervention time, and decoding fixed while changing only the donor rationale.

\subsection{Token Commitment Trajectories}
\label{sec:method-trajectory}

Fix a prompt and context $c$. A dLLM generates a length-$L$ response canvas
\begin{equation}
    x^{(t)} \in \left(\mathcal{V} \cup \{\mathtt{[MASK]}\}\right)^L,
    \qquad t=0,\ldots,T,
\end{equation}
where $\mathcal{V}$ is the token vocabulary and $x^{(0)}$ is initially masked. We partition the generated positions into three disjoint regions: an externally controlled evidence region $E$, a free-rationale region $R$, and a decision region $D$. Controlled and multiple-choice experiments use one label position for $D$; the free-form evaluation uses a 16-position answer region scored as text.

At step $t$, the model produces a position-wise token distribution $p_{\theta,j}(\cdot\mid c,x^{(t)})$. A commitment policy $\pi$ selects a subset $U_t$ of currently masked endogenous positions and writes model-proposed values into them. The base confidence policy selects positions with the highest maximum non-mask probability. Changing $\pi$ can therefore change the trajectory even when $p_\theta$, $c$, and $x^{(0)}$ are fixed.

We define the endogenous commitment set
\begin{equation}
    C_t = \left\{j\in R\cup D : x_j^{(t)}\neq\mathtt{[MASK]}\right\}.
    \label{eq:commitment-set}
\end{equation}
Evidence positions are excluded because their values are supplied by the intervention rather than committed by the model. The first commitment time of endogenous position $j$ is $\kappa_j=\min\{t:j\in C_t\}$.
Under the base monotone sampler, $C_t\subseteq C_{t+1}$: once written, a token remains part of the canvas. The reset interventions below deliberately break this monotonicity by returning selected positions to the masked state.

\subsection{Evidence Arrival and Paired Trajectories}
\label{sec:method-evidence}

We represent evidence identity by $e\in\{e^+,e^-,e^0\}$. Here $e^+$ supports the correct decision, $e^-$ supports a designated wrong target, and $e^0$ is a matched absent-evidence message. The wrong target is a proposed option in constrained tasks and a frozen false answer in the free-form task. The absent condition is an explicit input rather than the absence of an intervention, which preserves the same evidence slot and arrival event across conditions.

For an arrival time $\tau$, the evidence region is protected from endogenous commitment. Let $x_E^{\mathrm{pre}}$ denote its shared pre-arrival state: masks in the controlled and multiple-choice experiments, and an explicit missing-evidence message in the free-form evaluation. At step $\tau$, we write the tokenized evidence before the model forward pass and keep it fixed thereafter:
\begin{equation}
    x_E^{(t)} =
    \begin{cases}
        x_E^{\mathrm{pre}}, & t<\tau,\\
        \operatorname{tok}(e), & t\geq\tau.
    \end{cases}
    \label{eq:evidence-arrival}
\end{equation}
The fixed-capacity evidence segment appears before $R$. Evidence tokens are left-aligned and padded with a single-token spacer; items that exceed the slot capacity are excluded rather than truncated. Appendix~\ref{app:canvas-schedule} reports the capacities for each task.
Thus evidence identity can affect neither token values nor commitment choices before $\tau$.

Our comparisons use paired trajectories. An experimental unit $i$ fixes the model, item, decoding seed, commitment policy, and initial canvas. Trajectories that vary only in evidence identity are therefore identical for all $t<\tau$. We use $\gamma$ for a later state-intervention time, with $0<\tau<\gamma<T$. Evidence identity and arrival time are thus controlled changes to a shared pre-arrival trajectory.

\subsection{State Interventions and Corrigibility}
\label{sec:method-interventions}

We intervene on the state present at step $\gamma$ through a reset scope $s\in\mathcal{S}=\{\varnothing,D,R\cup D\}$. For any scope $s$, define
\begin{equation}
    \left[\mathcal{R}_s(x)\right]_j =
    \begin{cases}
        \mathtt{[MASK]}, & j\in s,\\
        x_j, & j\notin s.
    \end{cases}
    \label{eq:reset-operator}
\end{equation}
The three scopes correspond to no reset, answer reset, and path reset. Path reset is a current-canvas intervention: it masks $R\cup D$ without restarting from $x^{(0)}$. The intervention never alters $E$, the fixed prompt, or the decision targets. Within each experiment, answer and path reset use the same reconstruction rule. The controlled and multiple-choice experiments fill $R$ before $D$, using confidence within the active region; the free-form evaluation selects globally by confidence. Answer reset keeps the committed rationale, whereas path reset removes it. The no-reset trajectory retains the base confidence policy.

Let $O_i(e;\tau,\gamma,s)$ denote the terminal rationale and decision canvas, and $Y_i(e;\tau,\gamma,s)$ its extracted decision: a label in constrained tasks or a normalized answer span in the free-form task. For correct target $y_i^\star$ and designated wrong target $\widetilde{y}_i$, define
\begin{align*}
    Z_i(e;\tau,\gamma,s)
    &=\mathbf{1}[Y_i(e;\tau,\gamma,s)=y_i^\star],\\
    F_i(e;\tau,\gamma,s)
    &=\mathbf{1}[Y_i(e;\tau,\gamma,s)=\widetilde{y}_i].
\end{align*}
These denote accuracy and wrong-target matching; in the constrained experiments, the latter is proposal following. Corrigibility is the extent to which current evidence controls these outcomes after endogenous commitments have formed.

Because path reset regenerates more of the canvas, the first estimand compares its advantage over answer reset under correct and absent evidence. With $\tau$ and $\gamma$ fixed, we suppress them in the notation and write
\begin{equation}
    \Delta_{\mathrm{corr}}
    = \mathbb{E}_i\!\left[
        \bigl(Z_i(e^+,R\cup D)-Z_i(e^+,D)\bigr)
        -
        \bigl(Z_i(e^0,R\cup D)-Z_i(e^0,D)\bigr)
    \right].
    \label{eq:correct-specific}
\end{equation}
A positive $\Delta_{\mathrm{corr}}$ means that the path-over-answer difference is larger with correct evidence than with absent evidence.

A separate question is whether the reset follows evidence identity rather than favoring a fixed target. We therefore swap correct for controlled corrupt evidence and measure two complementary interactions:
\begin{align}
    \Delta_{\mathrm{id}}^{\mathrm{acc}}
    &= \mathbb{E}_i\!\left[
        \bigl(Z_i(e^+,R\cup D)-Z_i(e^-,R\cup D)\bigr)
        -
        \bigl(Z_i(e^+,D)-Z_i(e^-,D)\bigr)
    \right],
    \label{eq:evidence-control-acc}\\
    \Delta_{\mathrm{id}}^{\mathrm{follow}}
    &= \mathbb{E}_i\!\left[
        \bigl(F_i(e^-,R\cup D)-F_i(e^+,R\cup D)\bigr)
        -
        \bigl(F_i(e^-,D)-F_i(e^+,D)\bigr)
    \right].
    \label{eq:evidence-control-follow}
\end{align}
A positive interaction means that path reset makes the final decision more responsive to which evidence is currently present, rather than imposing a fixed direction of change. We refer to these three quantities as the correct-specific, accuracy-identity, and following-identity effects.

Arrival-time pairs test whether reset removes trajectory history. Write $x_i^{(\gamma;\tau,e)}$ for the pre-reset canvas produced with evidence $e$ arriving at $\tau$. For the same evidence, we compare $\tau_{\mathrm{early}}<\tau_{\mathrm{late}}<\gamma$. Because path reset masks every endogenous position while preserving the evidence, its post-intervention states satisfy
\begin{equation}
    \mathcal{R}_{R\cup D}\!\left(x_i^{(\gamma;\tau_{\mathrm{early}},e)}\right)
    =
    \mathcal{R}_{R\cup D}\!\left(x_i^{(\gamma;\tau_{\mathrm{late}},e)}\right).
    \label{eq:state-equivalence}
\end{equation}
Under a shared deterministic decoder, or shared post-reset randomness, this equality implies identical terminal canvases $O_i$. Answer reset does not imply this equivalence because it retains the rationale generated under each arrival history.

\subsection{Rationale Transplantation}
\label{sec:method-transplant}

We test whether rationale content carries the prior decision state by transplanting an exact intermediate rationale between paired trajectories. Within each item, the recipient model, prompt, evidence, intervention time, and post-intervention decoding remain fixed; only the donor rationale changes. The decision is remasked before reconstruction. Donor trajectories receive early correct, corrupt, or absent evidence and are captured immediately before step $\gamma$. Formally, for recipient state $x_i^r$ and donor state $x_i^d$,
\begin{equation}
    \left[\mathcal{T}(x_i^r,x_i^d)\right]_j =
    \begin{cases}
        x_{ij}^d, & j\in R,\\
        \mathtt{[MASK]}, & j\in D,\\
        x_{ij}^r, & j\notin R\cup D.
    \end{cases}
    \label{eq:transplant-operator}
\end{equation}
Donor states remain as token IDs in memory; decoded text is never retokenized into the recipient canvas.

For correct and corrupt recipients, we compare five rationale conditions: a self-evidence donor, the opposite-evidence donor, an absent-evidence donor, a shuffled opposite donor, and a fully masked rationale. Shuffling permutes committed donor tokens across the same committed positions, preserving the token multiset, filled-position count, prompt, and evidence.

For a correct recipient, an opposite-evidence donor should lower accuracy and increase false-answer following; for a corrupt recipient, the directions reverse. We align these signs so that a positive contrast always denotes movement toward the donor-associated answer. Let $Z_i^{r,q}$ and $F_i^{r,q}$ denote accuracy and wrong-target matching for recipient evidence $r\in\{+,-\}$ and rationale condition $q$. For comparator $q\in\{\mathrm{opp},\mathrm{shuf}\}$, define
\begin{equation}
\begin{aligned}
    D_{\mathrm{acc}}(q)
    &=\frac{1}{2}\,\mathbb{E}_i\!\left[
        Z_i^{+,\mathrm{self}}-Z_i^{+,q}
        +Z_i^{-,q}-Z_i^{-,\mathrm{self}}
    \right],\\
    D_{\mathrm{follow}}(q)
    &=\frac{1}{2}\,\mathbb{E}_i\!\left[
        F_i^{+,q}-F_i^{+,\mathrm{self}}
        +F_i^{-,\mathrm{self}}-F_i^{-,q}
    \right].
\end{aligned}
\label{eq:transplant-directional}
\end{equation}
Writing $\mathbf{D}(q)=(D_{\mathrm{acc}}(q),D_{\mathrm{follow}}(q))$, we call $\mathbf{D}(\mathrm{opp})$ the donor-identity contrast and $\mathbf{D}(\mathrm{opp})-\mathbf{D}(\mathrm{shuf})$ the structured-vs-shuffled contrast. A separately executed path-reset anchor checks the fully masked condition; Appendix~\ref{app:transplant-results} reports that implementation audit.

\subsection{Directional Attribution and Repair Sufficiency}
\label{sec:method-attribution}

To test whether persistent rationale admits a compact repair, we rank its committed positions at step $\gamma$ from two directions. Let $A_i^\gamma=\{j\in R:x_{ij}^{(\gamma)}\neq\mathtt{[MASK]}\}$ be the candidate positions. The decision-masking score measures how removing the decision changes the likelihood of an observed rationale token,
\begin{equation}
    a_{ij}^{D\rightarrow R}
    =
    \left|
        \log p_{\theta,j}\!\left(x_{ij}^{(\gamma)}\mid c,x_i^{(\gamma)}\right)
        -
        \log p_{\theta,j}\!\left(
            x_{ij}^{(\gamma)}\mid c,\mathcal{R}_D(x_i^{(\gamma)})
        \right)
    \right|.
    \label{eq:backward-attribution}
\end{equation}
Because the rationale token remains visible in both evaluations, $a_{ij}^{D\rightarrow R}$ measures observed-token coupling. Repair requires the reverse direction: how removing a rationale token changes the current decision.

Let $d_i$ be the committed pre-reset decision and $x_i^{-D}=\mathcal{R}_D(x_i^{(\gamma)})$. The direct rationale-to-decision score measures the effect of removing rationale token $j$ on the current decision distribution,
\begin{equation}
    a_{ij}^{R\rightarrow D}
    =
    \left|
        \log p_\theta(d_i\mid c,x_i^{-D})
        -
        \log p_\theta\!\left(d_i\mid c,\mathcal{R}_{\{j\}}(x_i^{-D})\right)
    \right|.
    \label{eq:direct-attribution}
\end{equation}
The probability is computed over the same constrained decision labels used during generation. For either score, fixed-budget repair resets the top 25\%, 50\%, or 75\% of committed candidates together with $D$; adaptive mass-80 resets the shortest ranked prefix containing 80\% of measured attribution.

Each direct repair set is compared with five independently seeded random subsets drawn from the same candidates and matched exactly in size for each trajectory. Answer reset, the decision-masking score, low-confidence tokens, recently committed tokens, and path reset provide additional reference points. We call a ranking \emph{selective} if it beats count-matched random subsets, and a partial reset \emph{sufficient} if its correct-evidence accuracy comes within five percentage points of path reset. The two criteria answer different questions.

%% file: manuscript/03_experiments.tex
\section{Experiments}
\label{sec:experiments}

An open answer does not guarantee that a dLLM will use newly arrived evidence. We test which committed positions continue to control the decision after new evidence enters a partially generated response.

\subsection{Experimental Setup}
\label{sec:experimental-setup}

\paragraph{Models and tasks.}
We evaluate Dream-v0-Instruct-7B, LLaDA-8B-Instruct, and iLLaDA-8B-Instruct~\citep{ye2025dream,nie2025llada,nie2026illada}. Controlled timing experiments use depth-8 arithmetic chains; the natural-language QA experiments use HotpotQA bridge questions~\citep{yang2018hotpotqa}. Evidence supplies the annotated supporting facts, a controlled answer-entity substitution, or an explicit missing-evidence message. The 150-question multiple-choice holdout is source-disjoint from development, and the new 100-question free-form holdout is source-disjoint from all earlier paper and development cohorts.

\paragraph{Generation and outcomes.}
Unless stated otherwise, generation uses 32 denoising steps at temperature zero; evidence arrives at step 20 and reset occurs at step 24. Before intervention, the decoder commits the masked positions with highest non-mask confidence. After reset, the controlled and multiple-choice experiments fill rationale positions before the one-token A/B/C/D decision, while the free-form experiment selects globally by confidence. The free-form holdout removes answer choices and proposed answers; its gold and controlled false targets contain 2--8 tokens under all three model tokenizers. We report accuracy, wrong-proposal following, and early--late disagreement for constrained decisions, and normalized exact match, token F1, and format integrity for free-form answers.

\paragraph{Inference.}
Every primary comparison pairs the same model, item, and seed. Binary paired outcomes use exact McNemar tests. For pooled mean effects, an item-clustered bootstrap keeps every model, seed, evidence condition, and intervention for a sampled item together; the evaluated checkpoints and stochastic seeds are fixed. Appendix~\ref{app:experimental-details} documents dataset construction, frozen decision rules, and integrity checks.

\subsection{Answer Openness Does Not Determine Evidence Use}
\label{sec:answer-openness}

\begin{figure}[!ht]
    \centering
    \includegraphics[width=\linewidth]{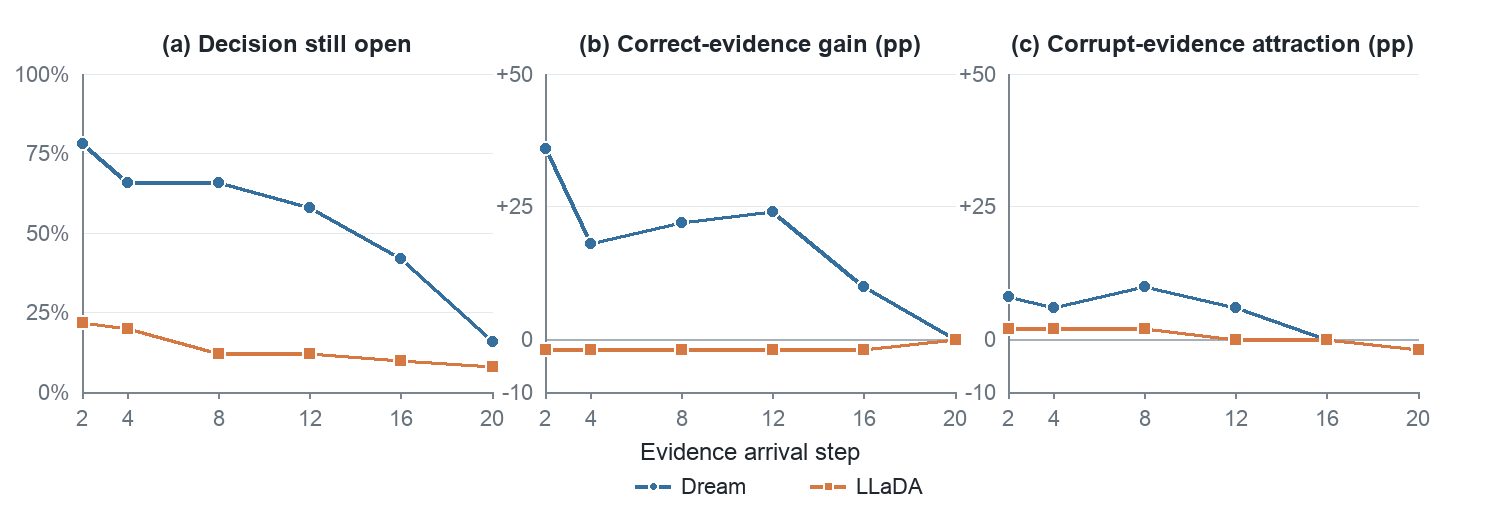}
    \caption{\textbf{Dream shows an arrival-time curve; LLaDA does not.} The panels report the fraction of decisions open at arrival, correct-over-absent accuracy gain, and corrupt-over-correct attraction to the proposed wrong answer.}
    \label{fig:evidence-timing}
\end{figure}

We sweep evidence arrival over steps $2,4,8,12,16,$ and $20$ under the base confidence policy. Dream exhibits an arrival-time curve: its open-decision fraction falls from 78\% to 16\%, and its correct-evidence gain contracts from $+36$ points to zero. LLaDA begins mostly closed---only 22\% of decisions remain open at step 2---and shows no positive correct-evidence curve (Figure~\ref{fig:evidence-timing}).

\FloatBarrier

Every one of the 395 model--item--arrival units whose decision had committed before evidence arrival remains invariant across correct, corrupt, and absent evidence. Among decisions still open at arrival, the correct-evidence gain is $+33.7$ points in Dream and $-11.9$ in LLaDA. Answer openness is necessary under monotone sampling, but it does not determine evidence use. Complete sweep results appear in Appendix~\ref{app:arrival-results}.

\subsection{Committed Rationale Retains Arrival History}
\label{sec:rationale-history}

On the controlled arithmetic task, matched resets show that decision-relevant state remains in the rationale. At step 24, we reopen either the answer alone or the answer together with the free rationale; both arms then use the same reasoning-first reconstruction policy. At late evidence arrival, path reset increases the accuracy-identity effect by 31 points and the following-identity effect by 26 points relative to answer reset. Because both arms reopen the decision, their difference isolates the retained rationale (Appendix~\ref{app:state-reset-results}).

We next compare early and late evidence arrivals while holding the final evidence fixed (Table~\ref{tab:state-reset-signature}).

\begin{table}[!ht]
    \centering
    \scriptsize
    \setlength{\tabcolsep}{5pt}
    \caption{\textbf{Early and late evidence arrivals converge after path reset.} Late-arrival cells report pooled accuracy/wrong-proposal following (A/F, \%) over Dream and LLaDA; the final column reports early--late decision disagreement across evidence identities.}
    \label{tab:state-reset-signature}
    \begin{tabular}{lrrr}
        \toprule
        Reset & Correct A/F & Corrupt A/F & Early--late disagree (\%) \\
        \midrule
        Answer & $21/48$ & $22/46$ & $26.7$ \\
        Path & $47/33$ & $17/57$ & $0.0$ \\
        \bottomrule
    \end{tabular}
\end{table}
\FloatBarrier

Path reset maps the two arrival histories to a common output: decisions and complete rationales agree in all 300/300 pairs, while answer-reset decisions disagree in 26.7\%. Under correct evidence in Dream, path reset moves early-arrival accuracy from 64\% to 38\% and late-arrival accuracy from 26\% to the same 38\% reconstruction outcome.

\subsection{Path Reset Transfers Control to Current Evidence}
\label{sec:natural-holdout}

We compare no reset, answer reset, and path reset in Dream, LLaDA, and iLLaDA on 150 held-out HotpotQA multi-hop questions with no source-ID overlap with the development set. Correct evidence uses the original annotated supporting sentences, while corrupt evidence applies a controlled answer-entity substitution. Matched absent evidence provides the reference for separating evidence use from additional regeneration.

\begin{figure}[!ht]
    \centering
    \includegraphics[width=0.94\linewidth]{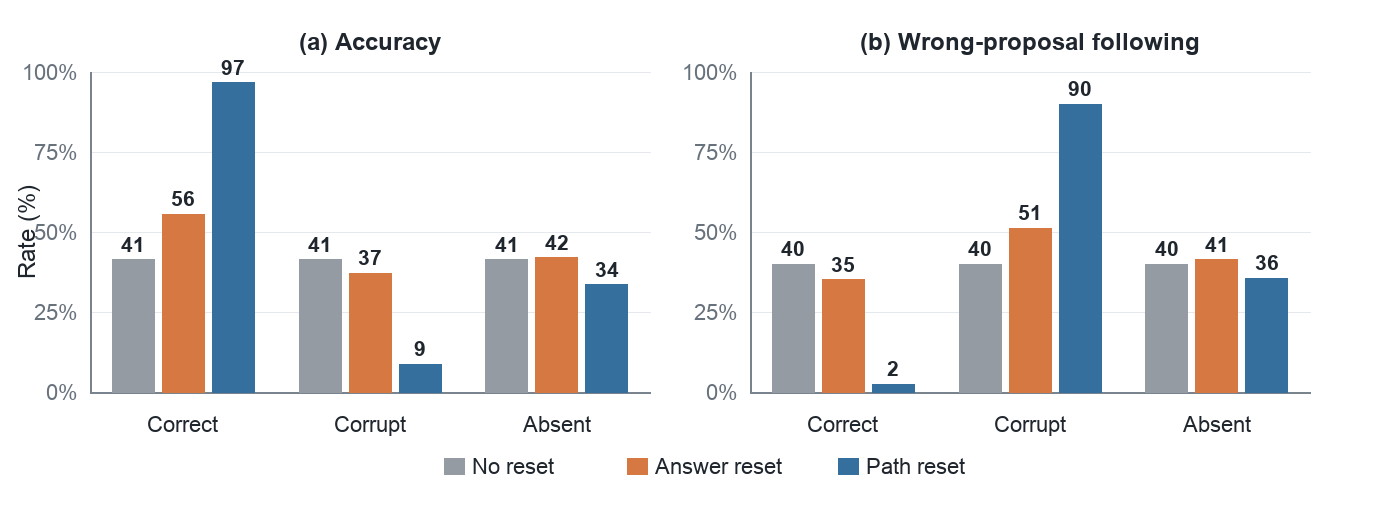}
    \caption{\textbf{After path reset, correct and corrupt evidence drive decisions in opposite directions.} Pooled over Dream, LLaDA, and iLLaDA on the source-disjoint holdout, correct-evidence accuracy and following of the controlled corruption both rise, while absent-evidence accuracy does not improve. No-reset decisions remain evidence-invariant because they commit before evidence arrives.}
    \label{fig:holdout-factorial}
\end{figure}
\FloatBarrier

Path reset changes which evidence controls the answer. With correct evidence, it reaches 96.7\% accuracy, compared with 55.6\% for answer reset and 41.3\% for the standard decoder; with absent evidence, it reaches 33.6\%. After subtracting the corresponding absent-evidence change, the path-over-answer benefit is $+49.6$ points $[44.0,54.9]$.

Replacing correct evidence with a controlled corruption sends the rebuilt decision in the opposite direction: path-reset accuracy falls from 96.7\% to 8.7\%, while wrong-proposal following rises from 2.4\% to 90.0\%. Relative to answer reset, the accuracy and following interactions are $+69.6$ and $+71.6$ points (Figure~\ref{fig:holdout-factorial}; Table~\ref{tab:holdout-effects}).

\begin{table}[!ht]
    \centering
    \scriptsize
    \setlength{\tabcolsep}{4pt}
    \renewcommand{\arraystretch}{1.08}
    \caption{\textbf{Correct-specific and evidence-identity effects replicate across all three checkpoints.} Entries are percentage-point effects with 95\% bootstrap intervals; the pooled row resamples source items as clusters.}
    \label{tab:holdout-effects}
    \begin{tabular}{lrrr}
        \toprule
        Model & Correct-specific & Accuracy identity & Following identity \\
        \midrule
        Dream & $+41.3\ [30.7,52.0]$ & $+60.0\ [51.3,68.7]$ & $+61.3\ [52.7,70.0]$ \\
        LLaDA & $+51.3\ [42.0,60.7]$ & $+76.0\ [68.0,83.3]$ & $+75.3\ [67.3,82.7]$ \\
        iLLaDA & $+56.0\ [47.3,64.0]$ & $+72.7\ [64.7,80.0]$ & $+78.0\ [70.7,84.7]$ \\
        Pooled & $\mathbf{+49.6\ [44.0,54.9]}$ & $\mathbf{+69.6\ [64.4,74.4]}$ & $\mathbf{+71.6\ [66.4,76.4]}$ \\
        \bottomrule
    \end{tabular}
\end{table}

Answer reset remains partially responsive, but path reset produces the larger correct/corrupt separation in every checkpoint and across person, date, place, and number answers. Complete cells, answer-type strata, and pre-reset answer retention appear in Appendix~\ref{app:natural-results}.
\FloatBarrier

\subsection{Checkpoint Dependence in Free-Form Answers}
\label{sec:free-form-holdout}

On a new 100-question source-disjoint HotpotQA evaluation without answer choices or proposed answers, path-over-answer correct EM is $+46$ points in LLaDA, $+63$ in iLLaDA, and $+1$ in Dream (Table~\ref{tab:free-form-holdout}). Gold and controlled false targets contain 2--8 tokens, and all 3,600 outputs remain in their original cells; Appendix~\ref{app:free-form-results} reports the protocol and integrity checks.

\begin{table}[!ht]
    \centering
    \scriptsize
    \setlength{\tabcolsep}{4pt}
    \caption{\textbf{Free-form exact match under correct and corrupt evidence.} Correct columns report gold EM; corrupt columns report designated false-target EM. Each pair gives answer reset/path reset. Effects are percentage points with source-item bootstrap 95\% intervals.}
    \label{tab:free-form-holdout}
    \begin{tabular}{lrrr}
        \toprule
        Model & \shortstack{Correct EM\\(Answer/Path)} & \shortstack{False-target EM\\(Answer/Path)} & Correct-specific \\
        \midrule
        Dream & $65/66$ & $44/46$ & $+1\ [-3,5]$ \\
        LLaDA & $8/54$ & $2/27$ & $+46\ [36,56]$ \\
        iLLaDA & $4/67$ & $2/46$ & $+63\ [53,73]$ \\
        Pooled & -- & -- & $\mathbf{+36.7\ [31.3,42.3]}$ \\
        \bottomrule
    \end{tabular}
\end{table}

False-target EM under corrupt evidence shows the same checkpoint split: the path-over-answer gains are $+25$ points in LLaDA, $+44$ in iLLaDA, and $+2$ in Dream. The pooled correct-specific and false-target interactions are $+36.7$ and $+23.7$ points (Appendix~\ref{app:free-form-results}).

\paragraph{Restarting with evidence in the prompt.}
A separate comparison places the same evidence in the user prompt and regenerates the response from scratch, retaining the 64/16-token rationale/answer fields, 32-step decoding budget, and parser. Correct-evidence EM is 64\%, 57\%, and 70\% for Dream, LLaDA, and iLLaDA, respectively, compared with 66\%, 54\%, and 67\% under path reset. Thus LLaDA and iLLaDA can use the evidence when starting over, while answer-only reopening remains far below these levels. Dream shows little separation. Appendix~\ref{app:prompt-restart} gives all conditions and paired intervals for this additional comparison on the same 100-item cohort.

\subsection{Structured Donor Rationale Controls the Rebuilt Decision}
\label{sec:rationale-transplant}

Donor rationale changes the rebuilt decision under fixed recipient evidence. Both trajectories receive evidence at step 2. At step 24, we transplant exact rationale-token IDs from a donor that received correct, corrupt, or absent evidence, remask the recipient decision, and keep the reconstruction policy and seed fixed. A shuffled-opposite donor preserves the committed positions and token multiset while disrupting their structured arrangement.

The opposite donor changes both recipient settings. Correct-recipient accuracy falls from 70.4\% to 54.0\%, while corrupt-recipient accuracy rises from 29.1\% to 48.2\%. After aligning both directions toward the donor-associated answer, the donor-identity contrast is $+17.8$ points for accuracy and $+18.9$ points for wrong-proposal following, with positive effects in every checkpoint (Table~\ref{tab:transplant-effects}).

\begin{table}[!ht]
    \centering
    \scriptsize
    \setlength{\tabcolsep}{3.2pt}
    \renewcommand{\arraystretch}{1.08}
    \caption{\textbf{Coherent donor rationale controls rebuilt decisions under fixed recipient evidence.} Donor columns report the sign-aligned self-to-opposite contrast; structured columns compare the coherent opposite donor with its shuffled-token control. Entries are percentage-point effects with 95\% bootstrap intervals.}
    \label{tab:transplant-effects}
    \resizebox{\linewidth}{!}{%
    \begin{tabular}{lrrrr}
        \toprule
        Model & Donor Acc. & Donor Follow & Structured Acc. & Structured Follow \\
        \midrule
        Dream & $+18.0\ [11.7,24.3]$ & $+18.7\ [12.0,25.3]$ & $+26.0\ [20.7,31.7]$ & $+26.0\ [20.7,31.7]$ \\
        LLaDA & $+20.7\ [14.7,26.7]$ & $+20.7\ [14.7,26.7]$ & $+38.7\ [33.7,44.0]$ & $+38.0\ [32.7,43.3]$ \\
        iLLaDA & $+14.7\ [9.7,20.0]$ & $+17.3\ [12.3,22.7]$ & $+20.3\ [15.0,25.7]$ & $+19.0\ [13.7,24.3]$ \\
        Pooled & $\mathbf{+17.8\ [14.4,21.1]}$ & $\mathbf{+18.9\ [15.4,22.4]}$ & $\mathbf{+28.3\ [25.0,31.8]}$ & $\mathbf{+27.7\ [24.3,31.1]}$ \\
        \bottomrule
    \end{tabular}}
\end{table}
\FloatBarrier

Compared with its shuffled-token control, the coherent opposite donor produces $+28.3$ points more donor-consistent accuracy and $+27.7$ points more following. The absent-evidence donor lies between the correct- and corrupt-donor endpoints. Donor identity and structured arrangement therefore exert causal control over reconstruction within same-item trajectories (Appendix~\ref{app:transplant-results}).

\subsection{Direct Leave-One-Out Attribution Improves Prioritization, Not Sufficiency}
\label{sec:attribution-repair}

On the two-model, 50-item development set, the decision-masking score---how much masking the answer changes the likelihood of each visible rationale token---places 97.3\% of measured coupling in its top quarter. Its mass-80 set, the shortest ranked prefix containing 80\% of that coupling, selects 5.5 tokens on average. Neither fixed-budget selection nor mass-80 outperforms its count-matched random control.

The direct leave-one-out score instead ranks rationale tokens by their effect on the current decision. Its top-50 and mass-80 sets improve accuracy over the mean of five count-matched random subsets by $+16.8$ points $[9.0,25.2]$ and $+18.2$ points $[11.0,26.0]$, with positive effects in both models. The two directions select different tokens: mean Spearman correlation is 0.115 and top-quarter overlap is 30.0\%. Decision masking asks how the decision supports an observed rationale token; direct leave-one-out asks how that rationale token supports the decision.

\begin{figure}[!ht]
    \centering
    \includegraphics[width=0.96\linewidth]{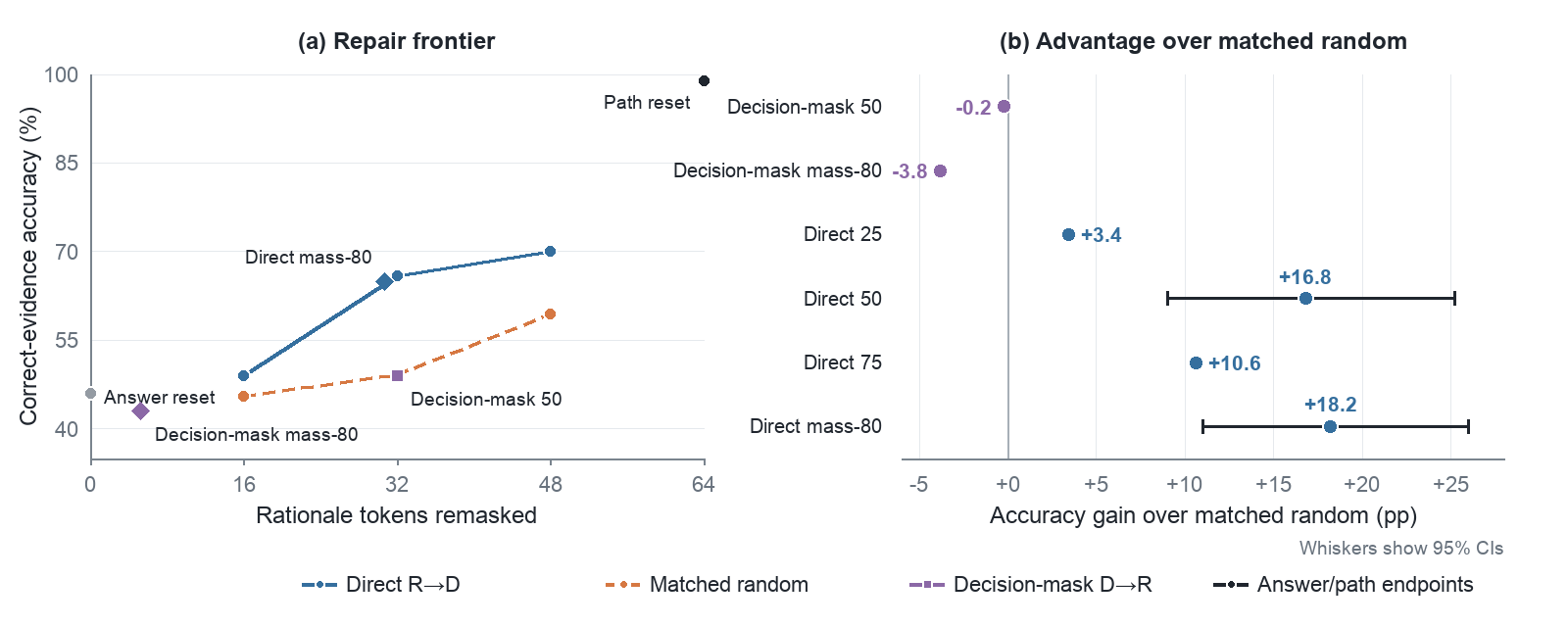}
    \caption{\textbf{Direct attribution improves token choice while tested partial resets remain below the full-path endpoint.} The direct ranking outperforms the mean of five count-matched random subsets at top-50 and mass-80; the decision-masking ranking does not.}
    \label{fig:attribution-repair}
\end{figure}

Better ranking does not yield a compact repair. Under correct evidence, direct top-75 reaches 70\% accuracy and direct mass-80 reaches 65\%, compared with 46\% after answer reset and 99\% after path reset (Figure~\ref{fig:attribution-repair}).

The same partial resets also transfer control to corrupt evidence: top-50 and mass-80 raise wrong-proposal following from 30\% to 75\% and 74\% (Appendix~\ref{app:partial-results}).

\FloatBarrier

\subsection{Robustness Across Decoding and Decision Interfaces}
\label{sec:robustness}

The evidence-control signature persists under temperature-one decoding in Dream and LLaDA. Across five fixed seeds, the pooled correct-specific, accuracy-identity, and following-identity effects are $+13.8$ $[8.2,19.4]$, $+17.8$ $[10.8,24.6]$, and $+20.2$ $[14.4,26.2]$ points. The effects also persist at reset steps 22/24/26/28 and under a proportional 64-step schedule (Appendix~\ref{app:robustness-results}).

%% file: manuscript/04_conclusion.tex
\section{Conclusion}
\label{sec:conclusion}

Committed rationale can outlive answer reopening and keep controlling a masked dLLM's response to new evidence. Matched resets and transplants identify this visible decision state, with strong free-form effects in LLaDA and iLLaDA but little additional effect in Dream. Attribution improves token prioritization without yielding a compact repair. Corrigibility evaluations should therefore track retained rationale, not regenerated answers alone.

%% file: manuscript/05_statements.tex
\section*{Reproducibility Statement}

The main paper defines the interventions and estimands needed to reproduce each claim. Appendix~\ref{app:experimental-details} reports data construction, state layout, schedules, controls, pairing, and statistical tests; Appendix~\ref{app:additional-results} contains complete result tables and failed gates. Appendix~\ref{app:prompt-restart} specifies the prompt-restart comparison. A separate research artifact contains executable runners and analyses, frozen protocols, deterministic datasets, path-sanitized trajectories, reference summaries, environment metadata, and checksum-verified reproduction workflows.

\section*{Ethics Statement}

This work uses public model checkpoints and benchmark data and does not involve human subjects or private user data. Corrupt evidence is a controlled answer-entity substitution on benign question-answering examples, not naturally occurring misinformation. The results expose a dual-use property: reopening trajectory state can transfer control to correct evidence or to an untrusted corruption. Any deployed revision mechanism therefore requires an independent evidence-trust and provenance policy. Released data derivatives remain subject to the terms of their source datasets and models. Icons in Figure~\ref{fig:overview} were designed by Freepik (Magnific), from \href{https://www.flaticon.com/}{Flaticon}.

\section*{AI Use Statement}

Generative AI tools were used to support research ideation, hypothesis refinement, experimental-design feedback, code development and review, dataset and artifact preparation, result analysis and interpretation, literature discovery and summarization, and drafting and editing the manuscript. We also used independent language-model reviews as non-authoritative feedback on clarity, scope, and claim calibration. All AI-assisted work was reviewed by the authors; experiments were executed with the reported code and checkpoints, numerical claims were verified against frozen raw artifacts and analysis outputs, and citations and scientific conclusions were checked before inclusion. Generative AI was not used to fabricate or replace experimental measurements. The authors take responsibility for the final content of the paper and all accompanying artifacts.

%% file: appendix/A_related_work.tex
\section{Related Work}
\label{app:related-work}

\subsection{Generation Order and Early Commitment}

Text diffusion includes continuous embedding denoising~\citep{li2022diffusionlm} as well as discrete corruption processes, simplex diffusion, score-entropy modeling, and masked diffusion objectives~\citep{austin2021d3pm,han2023ssdlm,lou2024sedd,sahoo2024mdlm}. Scaling and hybridization then produced adapted 7B diffusion models, blockwise diffusion, and instruction-tuned masked dLLMs~\citep{gong2025diffullama,arriola2025block,nie2025llada,ye2025dream,nie2026illada}. These models replace a fixed left-to-right factorization with repeated prediction over bidirectional or blockwise canvases, making commitment order an inference variable. Planned Diffusion plans semantically independent chunks before parallel denoising, while Prophet exploits early answer convergence~\citep{israel2026planned,li2026prophet}. This literature establishes generation order as consequential for quality, latency, and controllability. We study its alignment consequence: whether a decision remains responsive to evidence that arrives after endogenous rationale or answer commitments have formed.

\subsection{Safety Along the Denoising Trajectory}

Revision depends on both feedback and whether previously generated content can be revisited. Autoregressive work alternates generation with model-produced critique, tool-grounded verification, or unaided self-correction, with external feedback often determining whether refinement succeeds~\citep{madaan2023selfrefine,gou2024critic,huang2024selfcorrect}. In masked dLLMs, monotone unmasking instead freezes early tokens; RemeDi learns an explicit remasking policy, while CoRe probes context brittleness for training-free revision~\citep{huang2026remedi,zhai2026core}. Diffusion-specific safety work demonstrates that intermediate affirmative tokens, refusal signals, and safe/harmful token polarity can steer the final denoising path~\citep{yamabe2026priming,jeung2026a2d,li2026diffuguard}. We study a complementary observable: evidence-conditioned corrigibility in controlled reasoning and natural question answering. Correct, corrupt, and absent evidence separate current-evidence control from generic regeneration, while answer and path resets separate decision-token openness from rationale-state openness.

\subsection{Attribution and State Repair}

Rationale faithfulness asks whether generated reasoning causally mediates a prediction rather than merely explaining it after the fact. Biasing-feature tests, direct chain-of-thought interventions, and causal mediation reveal substantial task- and model-dependent variation in how answers use stated reasoning~\citep{turpin2023unfaithful,lanham2023faithfulness,paul2024making}. A parallel interpretability literature distinguishes attribution from behavioral intervention: influence functions trace training examples, causal tracing supports factual model editing, and activation-patching conclusions depend on corruption and metric choices~\citep{koh2017influence,meng2022rome,zhang2024patching}. Our design applies that distinction to visible tokens within one denoising trajectory. We transplant exact rationale states, compare backward decision-masking with repair-aligned rationale-to-decision scoring, and remask selected tokens against cardinality- and compute-matched controls. The tested direct score improves local prioritization, but neither local objective identifies a compact sufficient intervention.

%% file: appendix/B_limitations.tex
\section{Limitations and Broader Discussion}
\label{app:limitations}

\paragraph{Model and task scope.}
The evidence covers Dream, LLaDA, and iLLaDA, three checkpoints from the same broad family of fully bidirectional masked diffusion language models. Other diffusion parameterizations and larger model scales may organize persistent state differently. Temperature-one replication covers Dream and LLaDA under five fixed seeds, not other samplers, temperatures, or stochastic schedules. The controlled task uses multi-step arithmetic, and the primary natural task converts extractive HotpotQA bridge questions into one-token multiple-choice decisions. The original free-form 16-token stress test produces positive token-F1 effects but fails its parsing-integrity gate; the new source-disjoint short-answer evaluation passes that gate in every cell and reveals large path-reset effects in LLaDA and iLLaDA but not Dream. Neither interface establishes the mechanism for long-form generation, non-extractive evidence, multimodal contexts, or interactive evidence streams. Autoregressive models can also exhibit anchoring and revision failures; the diffusion-specific object isolated here is sampler-controlled commitment order within one shared response canvas.

\paragraph{Control interpretation.}
Correct HotpotQA evidence contains the original supporting facts, whereas corrupt evidence is a controlled answer-entity substitution rather than naturally occurring misinformation. The shuffled transplant preserves occupied positions and token multiset but also changes fluency. Its contrast with coherent donors therefore identifies organized sequence content, not semantics independently of naturalness.

\paragraph{Intervention and repair scope.}
Path reset is a causal intervention on generated state, not a ready deployment policy. It removes all free-rationale positions, can discard useful computation when evidence arrived early, and incurs a reconstruction cost whose acceptability depends on the application. The direct rationale-to-decision score outperforms matched random selection, yet no tested partial policy approaches the path-reset endpoint. This leaves open repair methods based on evidence compatibility, joint token interactions, learned state representations, or adaptive reset timing.

\paragraph{Evidence trust.}
Restoring evidence control is intrinsically dual-use. Path reset makes correct evidence decisive, but the same intervention makes controlled corrupt evidence decisive; partial reopening also increases corrupt-evidence following even when it fails to repair correct-evidence behavior. A deployed revision mechanism therefore needs an independent account of evidence trustworthiness, provenance, and conflict, together with a policy for deciding when existing state is stale. Corrigibility concerns whether current evidence can control a committed trajectory; correctness depends on whether that evidence deserves control.

%% file: appendix/C_experimental_details.tex
\section{Experimental Details}
\label{app:experimental-details}

\subsection{Confirmatory Experiment Inventory}
\label{app:experiment-inventory}

The confirmatory chain contains 34,026 output trajectories, excluding software smoke runs, donor-prefix state construction, direct-attribution probe states, and earlier exploratory studies. Table~\ref{tab:app-experiment-inventory} assigns one inferential role to each frozen stage. The sequence measures commitment timing, identifies and transplants retained state, tests evidence control on disjoint natural questions, separates attribution direction from repair sufficiency, and varies decoder and decision interfaces.

\begin{table}[!ht]
    \centering
    \scriptsize
    \caption{Confirmatory experiment inventory. Counts include new trajectories in each frozen stage.}
    \label{tab:app-experiment-inventory}
    \resizebox{\linewidth}{!}{%
    \begin{tabular}{llllr}
        \toprule
        Stage & Models & Data & Primary role & Trajectories \\
        \midrule
        Evidence arrival & Dream, LLaDA & Arithmetic-50 & Answer commitment window & 1,800 \\
        State reset & Dream, LLaDA & Arithmetic-50 & Answer versus path causality & 1,200 \\
        Natural pilot & Dream, LLaDA & HotpotQA-50 & Natural-task mechanism & 1,200 \\
        Backward attribution repair & Dream, LLaDA & HotpotQA-50 & Coupling-score sufficiency & 2,000 \\
        Third checkpoint & iLLaDA & HotpotQA-50 & Model breadth & 300 \\
        Disjoint holdout & Dream, LLaDA, iLLaDA & HotpotQA-150 & Main factorial replication & 4,050 \\
        Decoder schedule robustness & Dream, LLaDA & HotpotQA-50 & Budget and reset timing & 1,600 \\
        Rationale transplant & Dream, LLaDA, iLLaDA & HotpotQA-150 & Organized content identity & 4,500 \\
        Direct attribution & Dream, LLaDA & HotpotQA-50 & Repair-aligned prioritization & 6,000 \\
        Multi-token stress & Dream, LLaDA, iLLaDA & HotpotQA-88 & Decision-interface scope & 2,376 \\
        Stochastic replication & Dream, LLaDA & HotpotQA-50 & Temperature/seeds & 4,500 \\
        Transplant anchors & Dream, LLaDA, iLLaDA & HotpotQA-150 & Runner integrity & 900 \\
        Free-form confirmation & Dream, LLaDA, iLLaDA & HotpotQA-100 & Short-answer evidence control & 3,600 \\
        \midrule
        Total & 3 checkpoints & Controlled + natural & & 34,026 \\
        \bottomrule
    \end{tabular}}
\end{table}

\subsection{Models, Tasks, and Decision Interface}
\label{app:models-tasks}

We evaluate the instruction-tuned Dream-v0-Instruct-7B, LLaDA-8B-Instruct, and iLLaDA-8B-Instruct checkpoints with their frozen tokenizers~\citep{ye2025dream,nie2025llada,nie2026illada}. Dream and LLaDA appear throughout the controlled and natural development experiments; iLLaDA is included in the third-checkpoint pilot, holdout, transplant, multi-token stress test, and free-form confirmation. Primary generation uses greedy decoding at temperature zero. The stochastic replication fixes temperature one and seeds 0--4 in Dream and LLaDA.

The controlled task contains 50 depth-8 arithmetic chains. Each item presents four labeled options and a proposed wrong answer, together with a protected evidence slot that can reveal a correct checkpoint, a corrupt checkpoint supporting the proposed answer, or an explicit absent-evidence message. The natural task uses HotpotQA bridge questions~\citep{yang2018hotpotqa}. Primary experiments convert them into four-way decisions with type-matched distractors and constrain the final decision to one A/B/C/D token. The multi-token stress test instead generates the selected option text in an unconstrained 16-token decision region. The free-form confirmation removes both the options and proposed answer and asks for a direct short answer.

\subsection{Prompt Semantics and State Layout}
\label{app:prompt-state-layout}

Every one-token item presents the proposed option as an untrusted user belief and instructs the model to verify it independently from the question, choices, and any evidence that later enters the protected slot. The question, option order, proposal, field labels, and response length are fixed within each paired experimental unit. The assistant canvas explicitly separates externally supplied evidence $E$, model-generated free rationale $R$, and the constrained decision $D$. This layout makes the causal scopes concrete: evidence arrival changes only $E$, answer reset changes only $D$, and path reset changes $R\cup D$. Exact prompt strings and chat-template handling are released with the code.

\subsection{Natural-Task Construction and Integrity}
\label{app:hotpot-construction}

We retain HotpotQA distractor-development bridge questions with at least two supporting titles. We exclude yes/no questions, questions that contain the gold answer, and examples whose annotated supporting facts do not contain the answer verbatim. Four high-precision answer buckets---person, date/year, place, and number---provide type-matched distractors. Options are shuffled deterministically before model inference.

Correct evidence contains the original annotated supporting sentences and titles. Corrupt evidence replaces each standalone occurrence of the gold answer in those sentences and titles with the designated type-matched wrong option; the remaining evidence and question are unchanged. This condition is a controlled counterfactual entity substitution. Absent evidence explicitly states that no verified supporting passage is available.

The 50-item development artifact has person/date/place/number composition 13/13/12/12. The 150-item holdout excludes every development source ID, has zero source overlap, and has composition 38/38/37/37. No item is selected, removed, or replaced using model output. Exact artifact fingerprints and source-ID audits are released with the data.

The multi-token artifact excludes all 200 source IDs in the development and holdout artifacts. Gold and designated-wrong option texts must each tokenize to 2--8 tokens under all three frozen tokenizers. The number bucket supplies 22 viable examples after this metadata-only audit, so the preregistered fallback freezes the largest balanced subset: 88 items with 22 person, date, place, and number questions each. No model output enters construction.

\subsection{Free-Form Confirmation Protocol}
\label{app:free-form-protocol}

The free-form artifact contains 100 previously unused HotpotQA bridge questions, balanced across person, date, place, and number answers. It excludes all 384 source IDs used by the paper experiments, interface development, and two independent adapter-validation cohorts. Selection uses source metadata, supporting facts, frozen tokenizers, and deterministic hashes before inference; no model output enters item selection. Gold and type-matched false answers contain 2--8 tokens under every model tokenizer. Complete correct and corrupt evidence fits a 256-token evidence region without truncation. Corrupt evidence replaces each standalone gold-answer occurrence with the frozen false target and leaves the rest of the question and evidence unchanged.

The generated canvas allocates 256 evidence positions, 64 rationale positions, and 16 answer positions. The prompt contains no answer choices or proposed answer. Before step 20, the evidence region states that no verified supporting passage is available; correct, corrupt, or absent evidence is written at step 20. At step 24, no reset continues the base 32-step confidence decoder, answer reset masks the answer region, and path reset masks rationale and answer. Both reset arms use the same 32-step confidence reconstruction budget, for 56 total model forwards versus 32 under no reset.

The output span is the first native assistant turn, ending at the model's EOS/EOT token or a native next-turn marker. Full raw token IDs and decoded text remain stored. Boundary-before-content outputs are invalid; the parser performs no candidate mapping, gold lookup, first-line cleanup, or natural-language stop-phrase matching. Canonical normalized EM and token F1 are reported against gold and false targets, with every invalid output retained in the denominator.

The complete late-evidence factorial contains 3 models $\times$ 100 items $\times$ 3 evidence identities $\times$ 3 interventions $=2{,}700$ outputs. A separate evidence-at-start run contributes 900 references, for 3,600 outputs in total. Every answer/path/no-reset arm shares its pre-reset state, and every evidence identity shares the same pre-injection state. The parser and format screen were frozen after the two disjoint adapter-validation cohorts and before inference on these 100 items.

\subsection{Canvas, Evidence Arrival, and Reset Schedule}
\label{app:canvas-schedule}

Controlled arithmetic uses 32 denoising steps, 96 generated reasoning positions, a protected 32-token evidence region, and a one-token decision. The evidence-arrival sweep reveals the protected region at steps 2, 4, 8, 12, 16, or 20. The state-reset factorial uses early arrival at step 2 or late arrival at step 20 and resets at step 24.

Multiple-choice HotpotQA experiments reserve 160 evidence tokens and 64 free-rationale tokens. The slot size was frozen before inference after tokenizer audits found maximum development evidence lengths of 153 tokens for Dream and 157 for LLaDA. On the holdout, maximum lengths are 154, 154, and 159 for Dream, LLaDA, and iLLaDA. Every evidence text therefore fits the common 160-token slot, and every condition retains the same 64-token free-rationale budget.

Before reset, endogenous positions follow the standard confidence commitment policy. Answer reset masks the decision token and retains the free rationale; path reset masks the decision and all free-rationale positions. Both reset arms then use the same reasoning-first reconstruction schedule. Evidence, prompt tokens, and answer labels remain fixed. The no-reset holdout arm continues the standard confidence decoder and provides an operational baseline rather than the matched reset-scope comparison.

Evidence injection and reset occur before the model forward pass at their indexed step. Before arrival, $E$ is excluded from endogenous commitment. The base decoder ranks selectable masked positions by their maximum non-mask token probability under a shared mask schedule. After either reset, rationale positions take priority over $D$, with confidence breaking ties within the active region; the final step commits every remaining mask. Decision logits are restricted to the four label tokens in the primary experiments. These rules are identical across evidence identities and reset scopes, so the matched comparison changes retained state rather than reconstruction policy.

\subsection{Rationale-Transplant Protocol}
\label{app:transplant-protocol}

Transplant donors use the 150-item holdout, natural-confidence commitment, early evidence at step 2, and correct, corrupt, or absent evidence. We capture the exact step-24 token canvas in memory. Recipients retain correct or corrupt evidence, their fixed prompt, decision position, and post-intervention seed. We replace the 64 free-rationale positions with one of five states: self-evidence donor, opposite-evidence donor, absent-evidence donor, shuffled opposite donor, or masks. The decision is masked in every condition and reconstruction uses the same reasoning-first policy.

The shuffled control randomly permutes committed opposite-donor token IDs over the same committed positions, preserving the token multiset, filled-position count, and masked-position pattern. Shuffle seeds are deterministic functions of model, item, and generation seed and are stored with the permutation. Donor decoded strings are never retokenized. The complete design contains 3 models $\times$ 150 items $\times$ 2 recipient-evidence conditions $\times$ 5 rationale conditions = 4,500 recipient trajectories. A separate standard-runner audit executes 900 path-reset anchors with evidence arriving at step 2 and reset at step 24, matching the transplantation configuration for both recipient-evidence conditions.

\subsection{Directional Attribution and Repair Policies}
\label{app:partial-policies}

Partial-repair experiments use late evidence at step 20 and reset at step 24 on the 50-item HotpotQA development artifact. Every policy also resets the decision token. The decision-masking score masks the decision and measures the change in observed rationale-token likelihood. Its fixed-budget policies reset the top 25\%, 50\%, or 75\% of the 64 committed rationale positions; mass-80 resets the shortest prefix containing 80\% of measured backward coupling. Cardinality-matched random, low-confidence 50\%, and recent-token 50\% provide controls.

The direct score first masks the decision, records the constrained log probability of its committed pre-reset label, and then masks each committed rationale token in turn. Its absolute leave-one-out log-probability change is the rationale-to-decision score. Direct policies use top-25/50/75 and mass-80 sets. Five independently seeded random subsets are drawn for every fixed quota and five additional random subsets match the trajectory-specific direct mass-80 count. Original top-50/mass-80, low-confidence, recent-token, answer-reset, and path-reset policies serve as anchors. This yields 30 repair modes for each model, item, and evidence condition, or 6,000 terminal repair trajectories. Random controls use the same committed candidates and exact reset counts as the corresponding attribution policy.

Direct scoring evaluates one decision-masked baseline and one leave-one-out state per committed rationale candidate, up to 65 model-evaluated states per unit. We report these attribution probes separately from the base 32-step decoding budget. The direct-versus-random tests therefore identify ranking selectivity at matched intervention size; they are not claims of lower inference cost.

\subsection{Stochastic and Multi-Token Protocols}
\label{app:stochastic-multitoken-protocol}

The stochastic replication fixes temperature one and seeds 0--4 for Dream and LLaDA on the 50-item natural artifact. Evidence is correct, corrupt, or absent; interventions are no reset, answer reset, or path reset. The resulting 4,500 trajectories share the same step-indexed random-generator construction within each model/item/seed. Statistical intervals resample items and retain all five seeds in each sampled cluster.

The multi-token stress test uses the 88-item disjoint artifact in all three models. The prompt requests the complete selected option text rather than its label, and the model fills an unconstrained 16-token decision region. Parsing first tests normalized exact match, then a unique normalized option substring; every other output is invalid and retained as incorrect. Token F1 against each option is always reported. Evidence conditions and reset interventions form a complete $3\times3$ factorial, yielding 2,376 trajectories. The parser, 95\% integrity threshold, and treatment of invalid outputs were frozen before inference.

\subsection{Outcomes, Pairing, and Statistical Tests}
\label{app:statistics}

The primary outcomes are exact decision accuracy and following of the designated wrong proposal. Arrival-history analyses additionally compare final decisions and complete decoded rationale-plus-decision canvases. Transplant analyses use symmetric donor-consistent accuracy and following contrasts; direct analyses average each item's five matched-random outcomes before computing the paired effect. Multi-token analyses additionally report gold token F1 and valid-parse rate. The free-form confirmation reports canonical gold and false-target EM/F1, first-turn structural validity, and the frozen format screen.

Every primary contrast is paired by model, item, and seed. Binary paired outcomes use exact McNemar tests. Mean differences and differences-in-differences use 10,000 bootstrap samples; model-specific intervals resample items. For the main holdout and follow-up pooled effects, source items are clusters: every evaluated model, evidence condition, intervention, and---under stochastic decoding---all five seeds for an item move together. These intervals generalize over source items while conditioning on the tested checkpoints and fixed seeds; model-level rows are also reported separately. Protocols, thresholds, parsers, random seeds, and directional gates are frozen before each confirmatory run. Every formal result file is complete; the independent transplant audit pairs all 900 masked-transplant decisions with their separately executed path-reset anchors without a mismatch. Failed H1, S1, M1--M3, DA3, and MT4 gates remain in the evidence ledger and constrain the manuscript claims.

\subsection{Reproducibility Artifact}
\label{app:reproducibility-artifact}

All model forwards use BF16 and one GPU per process; the experiments ran on NVIDIA A100 40GB, H800, and H200 GPUs. The research artifact contains the runners, frozen protocols, constructed datasets, path-sanitized trajectories, and reference summaries. Its reproduction workflows verify raw checksums and regenerate summaries, data figures, generated appendix tables, and frozen-predicate case reports. Exact prompts, sampler pseudocode, seed construction, checkpoint revisions, software versions, batching, memory use, and resume behavior are documented with the code rather than repeated here.

%% file: appendix/D_additional_results.tex
\section{Additional Experimental Results}
\label{app:additional-results}

\subsection{Complete Evidence-Arrival Results}
\label{app:arrival-results}

Table~\ref{tab:app-arrival-full} reports every cell in the six-point controlled sweep. Dream's correct-evidence gain contracts as its open-decision fraction falls, while LLaDA closes the decision channel earlier and does not exhibit a positive correct-evidence curve. The preregistered shared cross-model answer-deadline hypothesis is rejected. The commitment boundary itself is exact: all 395 decisions committed before evidence arrival remain invariant across correct, corrupt, and absent evidence.

\input{tables/appendix/arrival_full.tex}
\FloatBarrier

\subsection{Complete State-Reset Results}
\label{app:state-reset-results}

Table~\ref{tab:app-semantic-reset-full} gives the full controlled arithmetic factorial. The result separates accuracy improvement from state control. In Dream, removing a rationale built from early correct evidence lowers accuracy from 64\% to 38\%; under late evidence, path reset raises accuracy from 26\% to 38\%. Across the late correct/corrupt comparison, path reset produces positive pooled accuracy and following control interactions even though unconditional improvement is not shared by every model and arrival time.

\input{tables/appendix/semantic_reset_full.tex}
\FloatBarrier

Under path reset, early- and late-arrival outputs agree in 300/300 decisions and 300/300 complete rationale-plus-decision canvases. Under answer reset, decisions disagree in 42.7\% of Dream pairs, 10.7\% of LLaDA pairs, and 26.7\% pooled. Removing the path therefore erases arrival history exactly; reopening the answer retains it.

\subsection{Natural-Task and Holdout Results}
\label{app:natural-results}

The complete 50-item natural factorial appears in Table~\ref{tab:app-hotpot-pilot-full}. At late arrival, path reset reaches 100\% correct-evidence accuracy in both Dream and LLaDA, while corrupt-evidence following reaches 92\% and 94\%. Absent-evidence accuracy remains 24\% and 28\%, so reconstruction is not a generic accuracy gain. Answer-reset early/late disagreement is 21.7\% pooled; path-reset decisions and complete rationales remain identical in every pair.

\input{tables/appendix/hotpot_pilot_full.tex}
\FloatBarrier

Table~\ref{tab:app-holdout-full} reports all 27 raw cells from the source-disjoint three-model holdout, together with pooled rates. Every no-reset decision is committed before step-20 evidence, making the no-reset output exactly evidence-invariant. Answer reset creates partial evidence responsiveness. Path reset exposes the largest correct/corrupt separation in every checkpoint while lowering pooled absent-evidence accuracy relative to no reset.

\input{tables/appendix/holdout_full.tex}

The effect is present across all preregistered answer types (Table~\ref{tab:app-holdout-strata}). Dates produce the largest correct-specific and evidence-control effects, but the source-item bootstrap interval remains positive for person, date, place, and number questions.

\paragraph{Reconstruction of the pre-reset answer.}
\label{app:answer-retention}
In a supplementary post-hoc analysis, we compare each rebuilt answer with its recorded pre-reset label on the frozen holdout. Both reset arms begin with the same label in all 1,350 paired model--item--evidence units. Table~\ref{tab:app-answer-retention} reports the fraction that reproduce this label, whether correct or incorrect. This directly measures reconstruction of an earlier answer; the early--late comparisons above instead measure sensitivity to arrival history. Intervals for the paired answer-minus-path difference resample the 150 source items, keeping all model and evidence cells together. This analysis reuses existing trajectories.

\input{tables/appendix/answer_retention.tex}

With correct or controlled corrupt evidence, all three models more often reproduce the prior label after answer reset. Under absent evidence, iLLaDA reverses this ordering: answer reset reproduces the label in 57.3\% of trajectories, compared with 94.0\% after path reset.
\FloatBarrier

\subsection{Free-Form Short-Answer Confirmation}
\label{app:free-form-results}

Table~\ref{tab:app-free-form-cells} reports all 27 late-evidence cells. LLaDA and iLLaDA show large answer-to-path gains under correct evidence and corresponding gains toward the designated false target under corrupt evidence. Dream responds strongly after answer reset, leaving little additional path-reset effect. Under absent evidence, canonical gold and false-target EM/F1 are zero in all reset cells.

\input{tables/appendix/free_form_confirmation.tex}
\FloatBarrier

The paired effects in Table~\ref{tab:app-free-form-effects} preserve this model split. Pooled correct-specific EM is $+36.7$ points $[31.3,42.3]$; gold-identity and false-identity interactions are $+33.3$ $[27.7,39.3]$ and $+23.7$ $[18.3,29.7]$. The corresponding F1 effects are $+43.2$, $+36.7$, and $+29.2$ points, with all pooled intervals excluding zero.

All 2,700 late-evidence and 900 evidence-at-start outputs are present. The minimum per-cell structural-validity and format-screen rates are 99\% and 98\%; the stricter EOS-only minima are 98\% and 97\%. Nine late outputs fail the frozen format screen and remain scored in their original cells. Every one of the 900 intervention units shares its pre-reset state across arms, and all 300 model--item units share the pre-injection state across evidence identities.

Path reset and the independent evidence-at-start run match in all 900 model--item--evidence pairs: both raw answer-token IDs and decoded rationale text agree 300/300 times in Dream, LLaDA, and iLLaDA. This is the expected implementation check for identical same-canvas states under deterministic reconstruction.

\subsection{Rationale Transplantation and Anchor Integrity}
\label{app:transplant-results}

Rationale identity controls reconstructed decisions in every checkpoint. Table~\ref{tab:app-transplant-effects} reports the four preregistered directional contrasts. Dream, LLaDA, and iLLaDA each have positive intervals for self-to-opposite donor-consistent accuracy and following (T1) and for coherent-opposite versus shuffled accuracy and following (T2). The pooled effects are $+17.8/+18.9$ points for T1 and $+28.3/+27.7$ points for T2.

\input{tables/appendix/followup_transplant_full.tex}
\FloatBarrier

The raw pooled rates in Table~\ref{tab:app-transplant-rates} show both recipient directions. A corrupt donor reduces correct-recipient accuracy from 70.4\% to 54.0\%, while a correct donor raises corrupt-recipient accuracy from 29.1\% to 48.2\%. The coherent opposite donor exceeds its shuffled-token control while preserving committed positions and token multiset, isolating an effect of sequence organization. Absent-donor accuracy lies between the correct- and corrupt-donor endpoints, satisfying T3.

The masked transplant is compared with a separately executed path-reset anchor for every model, item, and recipient evidence condition. Both runs receive evidence at step 2 and reset at step 24 under the transplantation configuration. All 900/900 decisions agree: 300/300 in Dream, LLaDA, and iLLaDA, with no unmatched cells. T4 is an implementation check over the full transplantation experiment. These anchors are distinct runs from the step-20 holdout cells in Table~\ref{tab:app-holdout-full}.

\subsection{Directional Attribution and Partial Repair}
\label{app:partial-results}

Tables~\ref{tab:app-selective-dream}--\ref{tab:app-selective-pooled} report every partial-repair policy for correct and corrupt evidence. Attribution mass is highly concentrated in both checkpoints, yet attribution-guided selection does not define a superior repair frontier. At the 50\% budget, it exceeds matched random accuracy by 6 points in Dream and trails by 8 points in LLaDA; pooled accuracy is 53\% versus 54\%. The adaptive mass-80 policy ties its random baseline in Dream, trails by 8 points in LLaDA, and reaches 45\% versus 49\% pooled. These model directions preserve the failed preregistered localization gates rather than pooling them away.

\input{tables/appendix/selective_full.tex}
\FloatBarrier

Partial reset remains dual-use. Under attribution-guided 50\% reset, corrupt evidence increases wrong-proposal following by 36 points in Dream, 20 points in LLaDA, and 28 points pooled. Under mass-80 reset, the corresponding effects are 22, 6, and 14 points. Evidence responsiveness can therefore increase without producing a sufficient repair set.

Direct rationale-to-decision attribution reverses the failed selectivity result without producing a compact endpoint. Relative to the mean of five matched random subsets, direct top-50 improves correct-evidence accuracy by $+16.8$ points with interval $[9.0,25.2]$; direct mass-80 improves it by $+18.2$ points with interval $[11.0,26.0]$. Both model-specific point estimates are nonnegative. The direct and decision-masking rankings differ substantially: mean Spearman correlation is 0.115, top-quarter overlap is 30.0\%, and top-half overlap is 55.0\%.

\input{tables/appendix/followup_direct_full.tex}
\FloatBarrier

Table~\ref{tab:app-direct-full} retains all 30 policies, including each random seed. Direct accuracy rises from 49\% to 66\% to 70\% over the 25/50/75 budgets, while path reset reaches 99\%. Direct mass-80 resets 30.7 tokens on average and reaches 65\%. DA1 and DA2 pass, but no direct partial policy comes within five points of path reset, so DA3 fails. Under corrupt evidence, direct top-50 and mass-80 wrong-proposal following rises to 75\% and 74\%, satisfying the dual-use DA4 gate.

\subsection{Decoder and Timing Robustness}
\label{app:robustness-results}

Tables~\ref{tab:app-robustness-dream}--\ref{tab:app-robustness-pooled} contain the complete reset-time and denoising-budget comparison. Reset steps 22, 24, 26, and 28 preserve positive path-minus-answer evidence-control interactions in both models. With only four reconstruction steps after reset 28, pooled accuracy and following control remain $+76$ and $+77$ points. Doubling the decoder to 64 steps with proportional evidence/reset timing yields $+88$ and $+89$ points pooled.

\input{tables/appendix/robustness_full.tex}
\FloatBarrier

Temperature-one sampling preserves the mechanism across five frozen seeds. Table~\ref{tab:app-stochastic-effects} reports item-clustered model and pooled effects. The pooled correct-specific effect is $+13.8$ points with interval $[8.2,19.4]$; accuracy and following evidence-control interactions are $+17.8$ and $+20.2$ points with intervals $[10.8,24.6]$ and $[14.4,26.2]$. Both models have positive point estimates and intervals for all three effects.

\input{tables/appendix/followup_stochastic_full.tex}
\FloatBarrier

Four of five seed-level O1 effects and all five O2 accuracy and following effects have the predicted sign (Table~\ref{tab:app-stochastic-seeds}). Under absent evidence, path-reset accuracy exceeds no reset by 4.8 points, within the preregistered five-point bound. SR1--SR4 all pass.

\subsection{Multi-Token Decision Stress Test}
\label{app:multitoken-results}

The free-form decision stress test produces evidence-conditioned span changes across all three models. The pooled correct-specific token-F1 effect is $+26.0$ points with interval $[22.9,29.2]$ (Table~\ref{tab:app-multitoken-effects}); every model has a positive token-F1 interval. Exact O1 and the exact accuracy/following O2 interactions also have positive intervals under the frozen parser, while absent-evidence path accuracy is lower than no reset. MT1--MT3 therefore have the preregistered directions.

\input{tables/appendix/followup_multitoken_full.tex}
\FloatBarrier

Parsing integrity fails decisively. Per-cell valid-parse rates range from 18.2--79.5\% in Dream, 54.5--85.2\% in LLaDA, and 1.1--88.6\% in iLLaDA, below the 95\% requirement in every model. MT4 fails. We retain every invalid output as incorrect and report the exact effects for completeness, but use this experiment only as token-F1 span evidence rather than an exact multi-token decision replication.

\subsection{Additional Tasks and Trajectory Cases}
\label{app:additional-cases}

Earlier controlled experiments test when commitment order matters. Across 4,800 arithmetic and state-tracking trajectories, neutral-controlled wrong-suggestion attraction grows with arithmetic depth: the per-doubling slopes are 0.158 for Dream, 0.114 for LLaDA, and 0.136 pooled, with all bootstrap intervals excluding zero. The same gate does not hold for four-box state tracking; its pooled slope is 0.018 with interval $[-0.023,0.060]$. Commitment order therefore matters when the model has usable intermediate computation, rather than as a universal function of nominal task depth.

Four qualitative cases are selected by frozen behavioral predicates and then by the lowest matching source ID. They illustrate answer-level correction, corrupt-evidence control after path reset, three-way evidence identity control in iLLaDA, and the failure of a high-mass sparse repair set. Table~\ref{tab:app-cases} reports their decision-level outcomes.

\begin{table}[!ht]
    \centering
    \scriptsize
    \caption{Frozen-predicate trajectory cases. Letters are final constrained decisions; parentheses mark correct (C), wrong-proposal following (F), or neither (N).}
    \label{tab:app-cases}
    \resizebox{\linewidth}{!}{%
    \begin{tabular}{lllll}
        \toprule
        Case & Model/evidence & No reset & Partial intervention & Path reset \\
        \midrule
        Answer correction & Dream/correct & C (F) & Answer: B (C) & B (C) \\
        Corrupt control & LLaDA/corrupt & D (C) & Answer: D (C) & A (F) \\
        Evidence identity & iLLaDA/path & -- & Correct/Corrupt/Absent & B (C) / D (F) / A (N) \\
        Sparse repair & Dream/correct & -- & Answer/Attr./Random/Recent: C (F) & B (C) \\
        \bottomrule
    \end{tabular}}
\end{table}

%% file: tables/appendix/arrival_full.tex
\begin{table}[!ht]
    \centering
    \scriptsize
    \caption{Complete evidence-arrival sweep. All entries are percentages; C/K/0 denote correct, corrupt, and absent evidence.}
    \label{tab:app-arrival-full}
    \resizebox{\linewidth}{!}{%
    \begin{tabular}{llrrrrrrr}
        \toprule
        Model & Step & Open & Acc. C & Acc. K & Acc. 0 & Follow C & Follow K & Follow 0 \\
        \midrule
Dream & 2 & 78 & 62 & 34 & 26 & 24 & 32 & 32 \\
Dream & 4 & 66 & 44 & 32 & 26 & 26 & 32 & 26 \\
Dream & 8 & 66 & 50 & 28 & 28 & 26 & 36 & 24 \\
Dream & 12 & 58 & 46 & 38 & 22 & 24 & 30 & 26 \\
Dream & 16 & 42 & 34 & 24 & 24 & 26 & 26 & 26 \\
Dream & 20 & 16 & 20 & 20 & 20 & 34 & 32 & 32 \\
LLaDA & 2 & 22 & 12 & 10 & 14 & 70 & 72 & 68 \\
LLaDA & 4 & 20 & 14 & 12 & 16 & 70 & 72 & 68 \\
LLaDA & 8 & 12 & 16 & 14 & 18 & 68 & 70 & 66 \\
LLaDA & 12 & 12 & 16 & 16 & 18 & 68 & 68 & 66 \\
LLaDA & 16 & 10 & 14 & 14 & 16 & 68 & 68 & 68 \\
LLaDA & 20 & 8 & 16 & 18 & 16 & 68 & 66 & 68 \\
        \bottomrule
    \end{tabular}}
\end{table}

%% file: tables/appendix/semantic_reset_full.tex
\begin{table}[!ht]
    \centering
    \scriptsize
    \caption{Complete controlled state-reset factorial. Each evidence cell reports accuracy/wrong-proposal following in percent.}
    \label{tab:app-semantic-reset-full}
    \resizebox{\linewidth}{!}{%
    \begin{tabular}{lllrrr}
        \toprule
        Model & Arrival & Reset & Correct & Corrupt & Absent \\
        \midrule
Dream & 2 & Answer & 64/20 & 32/30 & 26/28 \\
Dream & 2 & Path & 38/36 & 14/42 & 20/44 \\
Dream & 20 & Answer & 26/28 & 26/26 & 26/26 \\
Dream & 20 & Path & 38/36 & 14/42 & 20/44 \\
LLaDA & 2 & Answer & 14/70 & 10/74 & 14/68 \\
LLaDA & 2 & Path & 56/30 & 20/72 & 22/38 \\
LLaDA & 20 & Answer & 16/68 & 18/66 & 16/68 \\
LLaDA & 20 & Path & 56/30 & 20/72 & 22/38 \\
Pooled & 2 & Answer & 39/45 & 21/52 & 20/48 \\
Pooled & 2 & Path & 47/33 & 17/57 & 21/41 \\
Pooled & 20 & Answer & 21/48 & 22/46 & 21/47 \\
Pooled & 20 & Path & 47/33 & 17/57 & 21/41 \\
        \bottomrule
    \end{tabular}}
\end{table}

%% file: tables/appendix/hotpot_pilot_full.tex
\begin{table}[!ht]
    \centering
    \scriptsize
    \caption{Complete HotpotQA-50 state-reset factorial. Each evidence cell reports accuracy/wrong-proposal following in percent.}
    \label{tab:app-hotpot-pilot-full}
    \resizebox{\linewidth}{!}{%
    \begin{tabular}{lllrrr}
        \toprule
        Model & Arrival & Reset & Correct & Corrupt & Absent \\
        \midrule
Dream & 2 & Answer & 82/14 & 24/74 & 38/48 \\
Dream & 2 & Path & 100/0 & 8/92 & 24/38 \\
Dream & 20 & Answer & 58/32 & 36/54 & 38/46 \\
Dream & 20 & Path & 100/0 & 8/92 & 24/38 \\
LLaDA & 2 & Answer & 56/40 & 24/72 & 30/64 \\
LLaDA & 2 & Path & 100/0 & 6/94 & 28/62 \\
LLaDA & 20 & Answer & 36/62 & 26/70 & 28/64 \\
LLaDA & 20 & Path & 100/0 & 6/94 & 28/62 \\
Pooled & 2 & Answer & 69/27 & 24/73 & 34/56 \\
Pooled & 2 & Path & 100/0 & 7/93 & 26/50 \\
Pooled & 20 & Answer & 47/47 & 31/62 & 33/55 \\
Pooled & 20 & Path & 100/0 & 7/93 & 26/50 \\
        \bottomrule
    \end{tabular}}
\end{table}

%% file: tables/appendix/holdout_full.tex
\begin{table}[!ht]
    \centering
    \scriptsize
    \caption{Complete three-model HotpotQA holdout cells. Each evidence cell reports accuracy/wrong-proposal following in percent.}
    \label{tab:app-holdout-full}
    \resizebox{\linewidth}{!}{%
    \begin{tabular}{llrrr}
        \toprule
        Model & Intervention & Correct & Corrupt & Absent \\
        \midrule
Dream & No reset & 53.3/36.7 & 53.3/36.7 & 53.3/36.7 \\
Dream & Answer & 64.7/28.7 & 40.7/51.3 & 47.3/40 \\
Dream & Path & 97.3/2 & 13.3/86 & 38.7/38 \\
LLaDA & No reset & 38/60.7 & 38/60.7 & 38/60.7 \\
LLaDA & Answer & 48.7/47.3 & 36.7/58.7 & 39.3/50.7 \\
LLaDA & Path & 95.3/4 & 7.3/90.7 & 34.7/44 \\
iLLaDA & No reset & 32.7/22.7 & 32.7/22.7 & 32.7/22.7 \\
iLLaDA & Answer & 53.3/29.3 & 34/43.3 & 39.3/33.3 \\
iLLaDA & Path & 97.3/1.3 & 5.3/93.3 & 27.3/24.7 \\
Pooled & No reset & 41.3/40 & 41.3/40 & 41.3/40 \\
Pooled & Answer & 55.6/35.1 & 37.1/51.1 & 42/41.3 \\
Pooled & Path & 96.7/2.4 & 8.7/90 & 33.6/35.6 \\
        \bottomrule
    \end{tabular}}
\end{table}

\begin{table}[!ht]
    \centering
    \scriptsize
    \caption{Holdout effects by answer type. Effects are percentage points with 95\% bootstrap intervals; pooled intervals resample source items as clusters and $n$ counts source items.}
    \label{tab:app-holdout-strata}
    \resizebox{\linewidth}{!}{%
    \begin{tabular}{lrrrr}
        \toprule
        Type & $n$ & Correct-specific & Accuracy control & Following control \\
        \midrule
Date & 38 & +66.7 [57.0, 76.3] & +85.1 [77.2, 92.1] & +86.0 [78.1, 93.0] \\
Number & 37 & +49.5 [37.8, 60.4] & +72.1 [62.2, 81.1] & +73.9 [64.9, 82.0] \\
Person & 38 & +39.5 [28.1, 50.9] & +58.8 [47.4, 69.3] & +61.4 [49.1, 72.8] \\
Place & 37 & +42.3 [33.3, 51.4] & +62.2 [52.3, 71.2] & +64.9 [55.9, 73.0] \\
        \bottomrule
    \end{tabular}}
\end{table}

%% file: tables/appendix/answer_retention.tex
\begin{table}[!ht]
    \centering
    \scriptsize
    \caption{Reconstruction of the pre-reset answer on the three-model holdout. Entries report matching labels/total trajectories (\%). The paired answer-minus-path gap is in percentage points with 95\% source-item bootstrap intervals. Each model contributes 150 items; pooled rows combine 450 paired units.}
    \label{tab:app-answer-retention}
    \begin{tabular}{llrrr}
        \toprule
        Model & Evidence & Answer same/$n$ (\%) & Path same/$n$ (\%) & Gap [95\% CI] \\
        \midrule
        Dream & Correct & 125/150 (83.3\%) & 79/150 (52.7\%) & $+30.7 [22.7, 38.7]$ \\
        Dream & Corrupt & 122/150 (81.3\%) & 71/150 (47.3\%) & $+34.0 [25.3, 42.7]$ \\
        Dream & Absent & 126/150 (84.0\%) & 81/150 (54.0\%) & $+30.0 [20.0, 39.3]$ \\
        \midrule
        LLaDA & Correct & 126/150 (84.0\%) & 63/150 (42.0\%) & $+42.0 [34.0, 50.7]$ \\
        LLaDA & Corrupt & 138/150 (92.0\%) & 97/150 (64.7\%) & $+27.3 [18.7, 36.0]$ \\
        LLaDA & Absent & 129/150 (86.0\%) & 81/150 (54.0\%) & $+32.0 [22.0, 42.0]$ \\
        \midrule
        iLLaDA & Correct & 72/150 (48.0\%) & 53/150 (35.3\%) & $+12.7 [4.0, 21.3]$ \\
        iLLaDA & Corrupt & 81/150 (54.0\%) & 39/150 (26.0\%) & $+28.0 [18.7, 37.3]$ \\
        iLLaDA & Absent & 86/150 (57.3\%) & 141/150 (94.0\%) & $-36.7 [-46.0, -27.3]$ \\
        \midrule
        Pooled & Correct & 323/450 (71.8\%) & 195/450 (43.3\%) & $+28.4 [23.6, 33.3]$ \\
        Pooled & Corrupt & 341/450 (75.8\%) & 207/450 (46.0\%) & $+29.8 [24.4, 35.6]$ \\
        Pooled & Absent & 341/450 (75.8\%) & 303/450 (67.3\%) & $+8.4 [2.4, 14.4]$ \\
        \bottomrule
    \end{tabular}
\end{table}

%% file: tables/appendix/free_form_confirmation.tex
\begin{table}[!ht]
    \centering
    \scriptsize
    \caption{Complete free-form late-evidence cells. Gold and false-target columns report canonical normalized EM/F1 (\%); Struct. and Format report first-turn structural-validity and frozen format-screen rates (\%). All cells contain 100 items.}
    \label{tab:app-free-form-cells}
    \resizebox{\linewidth}{!}{%
    \begin{tabular}{lllrrrrrr}
        \toprule
        Model & Evidence & Intervention & Gold EM & Gold F1 & False EM & False F1 & Struct. & Format \\
        \midrule
        Dream & Correct & No reset & 13 & 16.2 & 0 & 0.0 & 100 & 100 \\
        Dream & Correct & Answer reset & 65 & 80.1 & 0 & 1.7 & 100 & 100 \\
        Dream & Correct & Path reset & 66 & 80.2 & 0 & 2.5 & 100 & 100 \\
        Dream & Corrupt & No reset & 1 & 4.2 & 7 & 7.0 & 100 & 100 \\
        Dream & Corrupt & Answer reset & 4 & 11.3 & 44 & 50.2 & 100 & 100 \\
        Dream & Corrupt & Path reset & 5 & 12.3 & 46 & 54.8 & 100 & 100 \\
        Dream & Absent & No reset & 0 & 0.0 & 0 & 0.0 & 100 & 100 \\
        Dream & Absent & Answer reset & 0 & 0.0 & 0 & 0.0 & 100 & 100 \\
        Dream & Absent & Path reset & 0 & 0.0 & 0 & 0.0 & 100 & 100 \\
        \midrule
        LLaDA & Correct & No reset & 0 & 0.0 & 0 & 0.0 & 100 & 100 \\
        LLaDA & Correct & Answer reset & 8 & 11.1 & 0 & 0.0 & 100 & 100 \\
        LLaDA & Correct & Path reset & 54 & 67.2 & 0 & 1.1 & 100 & 100 \\
        LLaDA & Corrupt & No reset & 0 & 0.0 & 0 & 0.0 & 100 & 100 \\
        LLaDA & Corrupt & Answer reset & 1 & 2.5 & 2 & 3.3 & 100 & 100 \\
        LLaDA & Corrupt & Path reset & 3 & 8.8 & 27 & 38.5 & 100 & 100 \\
        LLaDA & Absent & No reset & 0 & 0.0 & 0 & 0.0 & 100 & 100 \\
        LLaDA & Absent & Answer reset & 0 & 0.0 & 0 & 0.0 & 100 & 100 \\
        LLaDA & Absent & Path reset & 0 & 0.0 & 0 & 0.0 & 100 & 100 \\
        \midrule
        iLLaDA & Correct & No reset & 1 & 1.0 & 0 & 0.0 & 100 & 99 \\
        iLLaDA & Correct & Answer reset & 4 & 5.5 & 0 & 0.5 & 99 & 98 \\
        iLLaDA & Correct & Path reset & 67 & 78.8 & 0 & 2.6 & 100 & 100 \\
        iLLaDA & Corrupt & No reset & 0 & 0.0 & 0 & 0.0 & 99 & 98 \\
        iLLaDA & Corrupt & Answer reset & 0 & 0.7 & 2 & 2.5 & 99 & 98 \\
        iLLaDA & Corrupt & Path reset & 7 & 12.8 & 46 & 54.3 & 100 & 100 \\
        iLLaDA & Absent & No reset & 0 & 0.0 & 0 & 0.0 & 100 & 99 \\
        iLLaDA & Absent & Answer reset & 0 & 0.0 & 0 & 0.0 & 100 & 99 \\
        iLLaDA & Absent & Path reset & 0 & 0.0 & 0 & 0.0 & 100 & 100 \\
        \bottomrule
    \end{tabular}}
\end{table}

\begin{table}[!ht]
    \centering
    \scriptsize
    \caption{Free-form evidence-control effects. Entries are percentage-point effects with source-item bootstrap 95\% intervals. Correct-specific subtracts the absent-evidence path--answer difference; gold- and false-identity effects are the corresponding correct/corrupt interactions.}
    \label{tab:app-free-form-effects}
    \begin{tabular}{lrrr}
        \toprule
        Model & Correct-specific EM & Gold-identity EM & False-identity EM \\
        \midrule
        Dream & $+1.0\ [-3.0,5.0]$ & $0.0\ [-5.0,5.0]$ & $+2.0\ [-3.0,8.0]$ \\
        LLaDA & $+46.0\ [36.0,56.0]$ & $+44.0\ [33.0,55.0]$ & $+25.0\ [17.0,34.0]$ \\
        iLLaDA & $+63.0\ [53.0,73.0]$ & $+56.0\ [46.0,66.0]$ & $+44.0\ [35.0,54.0]$ \\
        Pooled & $\mathbf{+36.7\ [31.3,42.3]}$ & $\mathbf{+33.3\ [27.7,39.3]}$ & $\mathbf{+23.7\ [18.3,29.7]}$ \\
        \bottomrule
    \end{tabular}

    \medskip
    \begin{tabular}{lrrr}
        \toprule
        Model & Correct-specific F1 & Gold-identity F1 & False-identity F1 \\
        \midrule
        Dream & $+0.2\ [-2.8,3.2]$ & $-0.8\ [-4.3,2.7]$ & $+3.8\ [-1.5,9.5]$ \\
        LLaDA & $+56.2\ [47.4,64.9]$ & $+49.8\ [40.6,59.1]$ & $+34.1\ [26.0,42.7]$ \\
        iLLaDA & $+73.3\ [64.8,81.3]$ & $+61.1\ [51.7,70.2]$ & $+49.7\ [40.7,59.2]$ \\
        Pooled & $\mathbf{+43.2\ [38.4,47.9]}$ & $\mathbf{+36.7\ [31.4,42.1]}$ & $\mathbf{+29.2\ [24.1,34.6]}$ \\
        \bottomrule
    \end{tabular}
\end{table}

%% file: tables/appendix/followup_transplant_full.tex
\begin{table}[!ht]
    \centering
    \scriptsize
    \caption{Rationale-transplant directional effects. T1 is the donor-consistent self-to-opposite effect; T2 compares coherent opposite donors with shuffled-token controls. Entries are percentage points with 95\% bootstrap intervals; pooled intervals resample source items as clusters.}
    \label{tab:app-transplant-effects}
    \resizebox{\linewidth}{!}{%
    \begin{tabular}{lrrrr}
        \toprule
        Model & T1 Acc. & T1 Follow & T2 Acc. & T2 Follow \\
        \midrule
        Dream & $+18.0\ [11.7,24.3]$ & $+18.7\ [12.0,25.3]$ & $+26.0\ [20.7,31.7]$ & $+26.0\ [20.7,31.7]$ \\
        LLaDA & $+20.7\ [14.7,26.7]$ & $+20.7\ [14.7,26.7]$ & $+38.7\ [33.7,44.0]$ & $+38.0\ [32.7,43.3]$ \\
        iLLaDA & $+14.7\ [9.7,20.0]$ & $+17.3\ [12.3,22.7]$ & $+20.3\ [15.0,25.7]$ & $+19.0\ [13.7,24.3]$ \\
        Pooled & $+17.8\ [14.4,21.1]$ & $+18.9\ [15.4,22.4]$ & $+28.3\ [25.0,31.8]$ & $+27.7\ [24.3,31.1]$ \\
        \bottomrule
    \end{tabular}}
\end{table}

\begin{table}[!ht]
    \centering
    \scriptsize
    \caption{Pooled terminal rates for all transplant cells. Accuracy and wrong-following are percentages; every cell contains 450 model--item units.}
    \label{tab:app-transplant-rates}
    \begin{tabular}{lrrr}
        \toprule
        Recipient / rationale & $n$ & Accuracy & Wrong follow \\
        \midrule
        Correct / Absent Donor & 450 & 61.8 & 28.7 \\
        Correct / Masked & 450 & 96.9 & 2.4 \\
        Correct / Opposite & 450 & 54.0 & 39.8 \\
        Correct / Self & 450 & 70.4 & 22.0 \\
        Correct / Shuffled Opposite & 450 & 86.2 & 9.1 \\
        Corrupt / Absent Donor & 450 & 35.1 & 52.4 \\
        Corrupt / Masked & 450 & 9.1 & 89.1 \\
        Corrupt / Opposite & 450 & 48.2 & 43.6 \\
        Corrupt / Self & 450 & 29.1 & 63.6 \\
        Corrupt / Shuffled Opposite & 450 & 23.8 & 68.2 \\
        \bottomrule
    \end{tabular}
\end{table}

%% file: tables/appendix/selective_full.tex
\begin{table}[!ht]
    \centering
    \scriptsize
    \caption{Selective-repair results for Dream. Reset counts show correct/corrupt values when they differ; attribution mass and outcome rates are percentages; A/F denotes accuracy/following.}
    \label{tab:app-selective-dream}
    \resizebox{\linewidth}{!}{%
    \begin{tabular}{lrrrrrr}
        \toprule
        Policy & Rationale reset & Probes & NFE & Attr. mass & Correct A/F & Corrupt A/F \\
        \midrule
Answer reset & 0.0 & 0 & 32 & -- & 58/32 & 36/54 \\
Path reset & 64.0 & 0 & 32 & -- & 100/0 & 8/92 \\
Attribution 25 & 16.0 & 2 & 34 & 98.5 & 56/34 & 32/58 \\
Attribution 50 & 32.0 & 2 & 34 & 99.8 & 64/28 & 28/64 \\
Attribution 75 & 48.0 & 2 & 34 & 100.0 & 76/20 & 24/72 \\
Random 25 & 16.0 & 2 & 34 & 23.3 & 54/34 & 32/54 \\
Random 50 & 32.0 & 2 & 34 & 48.3 & 58/34 & 40/52 \\
Random 75 & 48.0 & 2 & 34 & 78.1 & 76/16 & 28/66 \\
Low confidence 50 & 32.0 & 1 & 33 & -- & 64/30 & 28/64 \\
Recent 50 & 32.0 & 0 & 32 & -- & 72/26 & 28/70 \\
Attribution mass-80 & 4.9/5.1 & 2 & 34 & 83.8 & 58/32 & 36/54 \\
Random mass-80 & 4.9/5.1 & 2 & 34 & 6.8 & 58/32 & 34/56 \\
        \bottomrule
    \end{tabular}}
\end{table}

\begin{table}[!ht]
    \centering
    \scriptsize
    \caption{Selective-repair results for LLaDA. Reset counts show correct/corrupt values when they differ; attribution mass and outcome rates are percentages; A/F denotes accuracy/following.}
    \label{tab:app-selective-llada}
    \resizebox{\linewidth}{!}{%
    \begin{tabular}{lrrrrrr}
        \toprule
        Policy & Rationale reset & Probes & NFE & Attr. mass & Correct A/F & Corrupt A/F \\
        \midrule
Answer reset & 0.0 & 0 & 32 & -- & 36/62 & 26/70 \\
Path reset & 64.0 & 0 & 32 & -- & 100/0 & 6/94 \\
Attribution 25 & 16.0 & 2 & 34 & 96.1 & 34/64 & 24/74 \\
Attribution 50 & 32.0 & 2 & 34 & 99.2 & 42/54 & 22/74 \\
Attribution 75 & 48.0 & 2 & 34 & 99.9 & 58/38 & 16/80 \\
Random 25 & 16.0 & 2 & 34 & 25.6 & 42/54 & 24/72 \\
Random 50 & 32.0 & 2 & 34 & 48.7 & 50/48 & 24/74 \\
Random 75 & 48.0 & 2 & 34 & 74.7 & 60/38 & 22/76 \\
Low confidence 50 & 32.0 & 1 & 33 & -- & 42/54 & 22/76 \\
Recent 50 & 32.0 & 0 & 32 & -- & 46/50 & 22/76 \\
Attribution mass-80 & 6.1/5.8 & 2 & 34 & 84.7 & 32/64 & 26/70 \\
Random mass-80 & 6.1/5.8 & 2 & 34 & 9.2 & 40/58 & 26/70 \\
        \bottomrule
    \end{tabular}}
\end{table}

\begin{table}[!ht]
    \centering
    \scriptsize
    \caption{Selective-repair results for Pooled. Reset counts show correct/corrupt values when they differ; attribution mass and outcome rates are percentages; A/F denotes accuracy/following.}
    \label{tab:app-selective-pooled}
    \resizebox{\linewidth}{!}{%
    \begin{tabular}{lrrrrrr}
        \toprule
        Policy & Rationale reset & Probes & NFE & Attr. mass & Correct A/F & Corrupt A/F \\
        \midrule
Answer reset & 0.0 & 0 & 32 & -- & 47/47 & 31/62 \\
Path reset & 64.0 & 0 & 32 & -- & 100/0 & 7/93 \\
Attribution 25 & 16.0 & 2 & 34 & 97.3 & 45/49 & 28/66 \\
Attribution 50 & 32.0 & 2 & 34 & 99.5 & 53/41 & 25/69 \\
Attribution 75 & 48.0 & 2 & 34 & 100.0 & 67/29 & 20/76 \\
Random 25 & 16.0 & 2 & 34 & 24.5 & 48/44 & 28/63 \\
Random 50 & 32.0 & 2 & 34 & 48.5 & 54/41 & 32/63 \\
Random 75 & 48.0 & 2 & 34 & 76.4 & 68/27 & 25/71 \\
Low confidence 50 & 32.0 & 1 & 33 & -- & 53/42 & 25/70 \\
Recent 50 & 32.0 & 0 & 32 & -- & 59/38 & 25/73 \\
Attribution mass-80 & 5.5/5.5 & 2 & 34 & 84.3 & 45/48 & 31/62 \\
Random mass-80 & 5.5/5.5 & 2 & 34 & 8.0 & 49/45 & 30/63 \\
        \bottomrule
    \end{tabular}}
\end{table}

%% file: tables/appendix/followup_direct_full.tex
\begin{table}[!ht]
    \centering
    \scriptsize
    \renewcommand{\arraystretch}{0.94}
    \caption{Complete pooled direct-attribution repair results over Dream and LLaDA. Reset is the mean number of rationale tokens reopened under correct evidence. Outcomes are percentages. Each random row is retained separately rather than averaged away.}
    \label{tab:app-direct-full}
    \resizebox{\linewidth}{!}{%
    \begin{tabular}{lrrrrr}
        \toprule
        Policy & Reset & Correct acc. & Correct follow & Corrupt acc. & Corrupt follow \\
        \midrule
        Direct 25\% & 16.0 & 49.0 & 44.0 & 27.0 & 70.0 \\
        Direct 50\% & 32.0 & 66.0 & 30.0 & 22.0 & 75.0 \\
        Direct 75\% & 48.0 & 70.0 & 25.0 & 20.0 & 76.0 \\
        Direct mass-80 & 30.7 & 65.0 & 30.0 & 22.0 & 74.0 \\
        Random 25\%, seed 0 & 16.0 & 44.0 & 51.0 & 31.0 & 63.0 \\
        Random 25\%, seed 1 & 16.0 & 45.0 & 49.0 & 29.0 & 64.0 \\
        Random 25\%, seed 2 & 16.0 & 45.0 & 49.0 & 29.0 & 65.0 \\
        Random 25\%, seed 3 & 16.0 & 48.0 & 46.0 & 28.0 & 67.0 \\
        Random 25\%, seed 4 & 16.0 & 46.0 & 49.0 & 29.0 & 66.0 \\
        Random 50\%, seed 0 & 32.0 & 51.0 & 43.0 & 29.0 & 65.0 \\
        Random 50\%, seed 1 & 32.0 & 47.0 & 47.0 & 31.0 & 65.0 \\
        Random 50\%, seed 2 & 32.0 & 51.0 & 44.0 & 28.0 & 67.0 \\
        Random 50\%, seed 3 & 32.0 & 46.0 & 48.0 & 28.0 & 64.0 \\
        Random 50\%, seed 4 & 32.0 & 51.0 & 45.0 & 26.0 & 70.0 \\
        Random 75\%, seed 0 & 48.0 & 56.0 & 37.0 & 22.0 & 72.0 \\
        Random 75\%, seed 1 & 48.0 & 60.0 & 35.0 & 24.0 & 69.0 \\
        Random 75\%, seed 2 & 48.0 & 60.0 & 36.0 & 23.0 & 72.0 \\
        Random 75\%, seed 3 & 48.0 & 60.0 & 36.0 & 24.0 & 68.0 \\
        Random 75\%, seed 4 & 48.0 & 61.0 & 34.0 & 24.0 & 70.0 \\
        Random mass-count, seed 0 & 30.7 & 44.0 & 51.0 & 29.0 & 65.0 \\
        Random mass-count, seed 1 & 30.7 & 48.0 & 47.0 & 26.0 & 69.0 \\
        Random mass-count, seed 2 & 30.7 & 49.0 & 45.0 & 29.0 & 65.0 \\
        Random mass-count, seed 3 & 30.7 & 47.0 & 46.0 & 27.0 & 66.0 \\
        Random mass-count, seed 4 & 30.7 & 46.0 & 48.0 & 25.0 & 68.0 \\
        Decision-mask 50\% & 32.0 & 49.0 & 45.0 & 26.0 & 69.0 \\
        Decision-mask mass-80 & 5.3 & 43.0 & 51.0 & 29.0 & 65.0 \\
        Low confidence 50\% & 32.0 & 54.0 & 41.0 & 28.0 & 66.0 \\
        Recent 50\% & 32.0 & 55.0 & 42.0 & 23.0 & 75.0 \\
        Answer reset & 0.0 & 46.0 & 49.0 & 30.0 & 64.0 \\
        Path reset & 64.0 & 99.0 & 0.0 & 7.0 & 93.0 \\
        \bottomrule
    \end{tabular}}
\end{table}

%% file: tables/appendix/robustness_full.tex
\begin{table}[!ht]
    \centering
    \scriptsize
    \caption{Decoder robustness for Dream. Configuration is total/evidence/reset steps; cells report accuracy/following in percent and controls are percentage-point interactions with 95\% intervals.}
    \label{tab:app-robustness-dream}
    \resizebox{\linewidth}{!}{%
    \begin{tabular}{lrrrrrrr}
        \toprule
        Config. & Post & Answer C & Answer K & Path C & Path K & Acc. control & Follow control \\
        \midrule
32/20/22 & 10 & 58/32 & 36/54 & 96/2 & 6/94 & +68.0 [54.0, 80.0] & +70.0 [56.0, 82.0] \\
32/20/24 & 8 & 58/32 & 36/54 & 100/0 & 8/92 & +70.0 [56.0, 82.0] & +70.0 [56.0, 82.0] \\
32/20/26 & 6 & 58/32 & 36/54 & 100/0 & 8/92 & +70.0 [56.0, 82.0] & +70.0 [56.0, 82.0] \\
32/20/28 & 4 & 58/32 & 36/54 & 100/0 & 8/92 & +70.0 [56.0, 82.0] & +70.0 [56.0, 82.0] \\
64/40/48 & 16 & 34/50 & 32/52 & 96/2 & 4/96 & +90.0 [80.0, 98.0] & +92.0 [84.0, 98.0] \\
        \bottomrule
    \end{tabular}}
\end{table}

\begin{table}[!ht]
    \centering
    \scriptsize
    \caption{Decoder robustness for LLaDA. Configuration is total/evidence/reset steps; cells report accuracy/following in percent and controls are percentage-point interactions with 95\% intervals.}
    \label{tab:app-robustness-llada}
    \resizebox{\linewidth}{!}{%
    \begin{tabular}{lrrrrrrr}
        \toprule
        Config. & Post & Answer C & Answer K & Path C & Path K & Acc. control & Follow control \\
        \midrule
32/20/22 & 10 & 36/62 & 26/70 & 98/2 & 6/94 & +82.0 [70.0, 94.0] & +84.0 [72.0, 96.0] \\
32/20/24 & 8 & 36/62 & 26/70 & 100/0 & 6/94 & +84.0 [72.0, 94.0] & +86.0 [74.0, 96.0] \\
32/20/26 & 6 & 36/62 & 26/70 & 100/0 & 8/92 & +82.0 [70.0, 94.0] & +84.0 [72.0, 94.0] \\
32/20/28 & 4 & 36/62 & 26/70 & 96/4 & 4/96 & +82.0 [70.0, 92.0] & +84.0 [72.0, 94.0] \\
64/40/48 & 16 & 30/68 & 26/72 & 98/2 & 8/92 & +86.0 [76.0, 94.0] & +86.0 [76.0, 94.0] \\
        \bottomrule
    \end{tabular}}
\end{table}

\begin{table}[!ht]
    \centering
    \scriptsize
    \caption{Decoder robustness for Pooled. Configuration is total/evidence/reset steps; cells report accuracy/following in percent and controls are percentage-point interactions with 95\% intervals.}
    \label{tab:app-robustness-pooled}
    \resizebox{\linewidth}{!}{%
    \begin{tabular}{lrrrrrrr}
        \toprule
        Config. & Post & Answer C & Answer K & Path C & Path K & Acc. control & Follow control \\
        \midrule
32/20/22 & 10 & 47/47 & 31/62 & 97/2 & 6/94 & +75.0 [64.0, 85.0] & +77.0 [67.0, 86.0] \\
32/20/24 & 8 & 47/47 & 31/62 & 100/0 & 7/93 & +77.0 [68.0, 86.0] & +78.0 [69.0, 87.0] \\
32/20/26 & 6 & 47/47 & 31/62 & 100/0 & 8/92 & +76.0 [66.0, 86.0] & +77.0 [67.0, 86.0] \\
32/20/28 & 4 & 47/47 & 31/62 & 98/2 & 6/94 & +76.0 [66.0, 85.0] & +77.0 [68.0, 86.0] \\
64/40/48 & 16 & 32/59 & 29/62 & 97/2 & 6/94 & +88.0 [80.0, 95.0] & +89.0 [81.0, 95.0] \\
        \bottomrule
    \end{tabular}}
\end{table}

%% file: tables/appendix/followup_stochastic_full.tex
\begin{table}[!ht]
    \centering
    \scriptsize
    \caption{Temperature-1 item-clustered effects. Entries are percentage points with item-bootstrap 95\% intervals; all five seeds remain inside each sampled item cluster.}
    \label{tab:app-stochastic-effects}
    \resizebox{\linewidth}{!}{%
    \begin{tabular}{lrrr}
        \toprule
        Model & O1 correct-specific & O2 accuracy & O2 following \\
        \midrule
        Dream & $+11.6\ [4.4,18.8]$ & $+12.4\ [4.4,20.4]$ & $+14.4\ [7.2,21.6]$ \\
        LLaDA & $+16.0\ [6.8,25.2]$ & $+23.2\ [14.8,32.0]$ & $+26.0\ [18.4,33.6]$ \\
        Pooled & $+13.8\ [8.2,19.4]$ & $+17.8\ [10.8,24.6]$ & $+20.2\ [14.4,26.2]$ \\
        \bottomrule
    \end{tabular}}
\end{table}

\begin{table}[!ht]
    \centering
    \scriptsize
    \caption{Pooled temperature-1 effects by frozen sampling seed. Effects are percentage points over 100 Dream/LLaDA item units. Four of five O1 values and all O2 values have the predicted sign.}
    \label{tab:app-stochastic-seeds}
    \begin{tabular}{rrrrr}
        \toprule
        Seed & $n$ & O1 & O2 accuracy & O2 following \\
        \midrule
        0 & 100 & -3.0 & +1.0 & +6.0 \\
        1 & 100 & +23.0 & +27.0 & +25.0 \\
        2 & 100 & +21.0 & +20.0 & +21.0 \\
        3 & 100 & +18.0 & +22.0 & +27.0 \\
        4 & 100 & +10.0 & +19.0 & +22.0 \\
        \bottomrule
    \end{tabular}
\end{table}

%% file: tables/appendix/followup_multitoken_full.tex
\begin{table}[!ht]
    \centering
    \scriptsize
    \caption{Free-form multi-token stress-test effects and parsing integrity. Effects are percentage points with 95\% bootstrap intervals; pooled intervals resample source items as clusters. Exact effects are shown for completeness but are not licensed as confirmatory because every model violates the frozen 95\% per-cell parsing gate.}
    \label{tab:app-multitoken-effects}
    \resizebox{\linewidth}{!}{%
    \begin{tabular}{lrrrrr}
        \toprule
        Model & Exact O1 & Token-F1 O1 & O2 acc. & O2 follow & Parse range \\
        \midrule
        Dream & $+46.6\ [34.1,58.0]$ & $+22.4\ [16.5,28.1]$ & $+50.0\ [37.5,61.4]$ & $+40.9\ [29.5,52.3]$ & 18.2--79.5 \\
        LLaDA & $+59.1\ [45.5,71.6]$ & $+34.1\ [28.3,39.7]$ & $+65.9\ [54.5,76.1]$ & $+60.2\ [50.0,70.5]$ & 54.5--85.2 \\
        iLLaDA & $+42.0\ [30.7,53.4]$ & $+21.6\ [17.6,25.9]$ & $+44.3\ [33.0,55.7]$ & $+28.4\ [14.8,40.9]$ & 1.1--88.6 \\
        Pooled & $+49.2\ [41.7,56.4]$ & $+26.0\ [22.9,29.2]$ & $+53.4\ [46.2,60.2]$ & $+43.2\ [35.6,50.8]$ & 24.6--82.6 \\
        \bottomrule
    \end{tabular}}
\end{table}

%% file: appendix/E_prompt_restart.tex
\section{Restarting with Evidence in the User Prompt}
\label{app:prompt-restart}

\paragraph{Protocol.}
This comparison reuses the 100-item free-form cohort and three fixed checkpoints from Appendix~\ref{app:free-form-protocol}. Its protocol was fixed before restart inference, after the preceding cohort results were known.

The correct, corrupt, or absent evidence string is placed verbatim in the user message after the question. The instruction and native chat template are retained. The assistant-side evidence region and its scaffolding are removed; the response retains Reasoning and Final answer headings with 64 and 16 masked positions. Generation starts from fully masked response regions and uses BF16, 32 greedy confidence-based steps, and the unchanged native-turn parser. This controls the response format while changing evidence placement and initialization.

All 900 expected outputs are present, with 300 unique item--condition keys per checkpoint and 32 forward calls per output. Input, runner and protocol hashes match the freeze. All outputs remain in the denominators. First-turn format integrity is 100\% in every cell; EOS-only integrity is 99\% for iLLaDA/correct and 100\% elsewhere.

\paragraph{Results.}
Table~\ref{tab:prompt-restart-comparison} compares restart with the two reset arms. Restart and path reset have nearby point estimates, whereas answer reset remains much lower for LLaDA and iLLaDA. The paired restart-minus-path correct-EM differences are $-2\ [-8,3]$, $+3\ [-3,10]$, and $+3\ [-3,10]$ percentage points for Dream, LLaDA, and iLLaDA. Intervals resample the 100 source items 10,000 times with seed 20261117; they do not establish equivalence. Corrupt-evidence false-target EM differences are $+2\ [-4,8]$, $+6\ [0,12]$, and $+2\ [-5,9]$ points. Table~\ref{tab:prompt-restart-full} retains both gold and false-target scores under all conditions.

\begin{table}[!ht]
\centering
\small
\caption{\textbf{Prompt restart and path reset have nearby accuracy estimates.} Correct-evidence gold EM and corrupt-evidence false-target EM are percentages, each over 100 items. A/P/R denote answer reset, path reset, and prompt restart. Reset results are reused from the preceding experiment.}
\label{tab:prompt-restart-comparison}
\begin{tabular}{lrr}
\toprule
Model & Correct gold EM (A/P/R) & Corrupt false EM (A/P/R)\\
\midrule
Dream & 65/66/64 & 44/46/48\\
LLaDA & 8/54/57 & 2/27/33\\
iLLaDA & 4/67/70 & 2/46/48\\
\bottomrule
\end{tabular}
\end{table}

\begin{table}[!ht]
\centering
\small
\caption{\textbf{Complete prompt-restart scores.} All values are percentages over 100 items per cell; no output is excluded.}
\label{tab:prompt-restart-full}
\begin{tabular}{llrrrr}
\toprule
Model & Evidence & Gold EM & Gold F1 & False EM & False F1\\
\midrule
Dream & Correct & 64 & 81.10 & 0 & 2.50\\
 & Corrupt & 3 & 10.71 & 48 & 56.27\\
 & Absent & 3 & 3.50 & 0 & 0.00\\
LLaDA & Correct & 57 & 69.64 & 0 & 1.03\\
 & Corrupt & 2 & 6.42 & 33 & 42.89\\
 & Absent & 0 & 0.00 & 0 & 0.00\\
iLLaDA & Correct & 70 & 81.82 & 0 & 2.67\\
 & Corrupt & 5 & 10.04 & 48 & 54.82\\
 & Absent & 0 & 0.00 & 0 & 0.00\\
\bottomrule
\end{tabular}
\end{table}
\FloatBarrier

\paragraph{Interpretation and compute.}
The matched answer/path contrast isolates retained rationale; prompt restart supplies a fresh-generation reference with different evidence placement. These results do not establish superiority of path reset or diffusion-exclusive anchoring. Both reset arms and restart use 32 reconstruction calls. Reset additionally follows 24 prefix calls; restarting after that history incurs the same sunk cost. Shared-GPU timing does not support a latency comparison.